\documentclass{article}

\usepackage[nonatbib, final]{neurips_2021}

\usepackage[utf8]{inputenc} 
\usepackage[T1]{fontenc}    
\usepackage{hyperref}       
\usepackage{url}            
\usepackage{booktabs}       
\usepackage{amsfonts}       
\usepackage{nicefrac}       
\usepackage{microtype}      
\usepackage[dvipsnames]{xcolor}

\usepackage{amsmath}

\usepackage{graphicx}
\usepackage{subfigure}
\usepackage{multirow}
\usepackage{multicol}

\RequirePackage{algorithm}
\RequirePackage{algorithmic}

\newcommand{\FG}[1]{\textcolor{ForestGreen}{{#1}}}
\newcommand{\RD}[1]{\textcolor{RubineRed}{{#1}}}

\title{OSOA: One-Shot Online Adaptation of Deep Generative Models for Lossless Compression}

\author{%
  Chen Zhang$^{\dagger}$ \quad Shifeng Zhang$^{\dagger}$ \quad Fabio M. Carlucci\thanks{Work done during employment at Huawei Technologies R\&D UK} \quad Zhenguo Li$^{\dagger}$  \\
  \\
  $^{\dagger}$Huawei Noah's Ark Lab\\
  \texttt{\{chenzhang10, zhangshifeng4, li.zhenguo\}@huawei.com}
}

\begin{document}

\maketitle

\begin{abstract}
Explicit deep generative models (DGMs), e.g., VAEs and Normalizing Flows, have shown to offer an effective data modelling alternative for lossless compression. 
However, DGMs themselves normally require large storage space and thus contaminate the advantage brought by accurate data density estimation.
To eliminate the requirement of saving separate models for different target datasets, we propose a novel setting that starts from a pretrained deep generative model and compresses the data batches while adapting the model with a dynamical system for only one epoch.
We formalise this setting as that of One-Shot Online Adaptation (OSOA) of DGMs for lossless compression and propose a vanilla algorithm under this setting. 
Experimental results show that vanilla OSOA can save significant time versus training bespoke models and space versus using one model for all targets.
With the same adaptation step number or adaptation time, it is shown vanilla OSOA can exhibit better space efficiency, e.g., $47\%$ less space, than fine-tuning the pretrained model and saving the fine-tuned model.
Moreover, we showcase the potential of OSOA and motivate more sophisticated OSOA algorithms by showing further space or time efficiency with multiple updates per batch and early stopping.
\end{abstract}

\section{Introduction}\label{sec:intro}
Lossless compression has always been an important part in the information and communications technology industry and has become more and more essential with the data explosion of the Big Data era~\cite{gage1994new,rabbani2002jpeg2000,roelofs1999png,collet2016smaller}.
Recently, a large number of machine learning approaches, predominantly deep generative models (DGMs), have been proposed for lossless compression~\cite{mentzer2019practical,townsend2019practical,hoogeboom2019integer,ho2019compression,berg2020idf++}.
As indicated by Shannon's source coding theorem, the compression limit is determined by the entropy of the data's ground truth distribution~\cite{mackay2003information}; any mismatch between an approximate distribution for coding and the unknown ground truth distribution will cause extra coding cost.
It is the powerful approximation ability of DGMs that enables more precise estimation of the ground truth distribution~\cite{kingma2013auto,salimans2017pixelcnn++,kingma2018glow,kumar2019videoflow} and enables DGMs based compression algorithms to outperform the conventional counter-parties in terms of the compression ratio.

However, in real production scenarios, both space and time efficiencies are desirable.
While more space efficiency is provided by higher compression ratio with DGMs based lossless compression algorithms, training of the probabilistic models is time consuming.
Training a separate model for each different dataset to compress would scale the expensive time cost linearly.
Moreover, to decode the dataset, one has to store the probabilistic model together with the codeword of the data, which requires additional space for local decoding and additional transmission cost for remote decoding.
Apart from one model per dataset, another extreme on the spectrum is using one probabilistic model for many different datasets.
Since each dataset can follow a very different distribution, using one model for all the datasets could lead to high inefficiency in terms of compression ratio.

To solve the above dilemma, we propose the setting of One-Shot Online Adaptation (OSOA) of deep generative models for lossless compression.
A DGM based lossless compression algorithm under the OSOA setting (referred to as OSOA algorithm) includes one pretrained deep generative model (referred to as base model) and one deterministic dynamical system.
The base model should be able to capture essential structures of a class of datasets to serve as a good starting point of the adaptation process.
The dynamical system is the mechanism that can update the current distribution to a new one based on the current batch at each step in the OSOA process.
At each OSOA step, one first uses the current deep generative model to evaluate the probabilistic mass function values to compress the current data batch before using the same batch to update the model with the dynamical system, which gives the name of online adaptation.
The online adaptation is only conducted for one epoch and a distribution sequence will be unrolled during this process.
Moreover, the dynamical system should be designed to optimise the objective functional associated with the deep generative model so that the sequence could provide better approximations on the unrolling procedure.
Under this new setting, we can start from a proper base model and benefit from improved compression ratio as the sequence being unrolled.
As the sequence is fully determined by the base model and the dynamical system, this approach allows to not only eliminate the need of extensively training a new model on each new dataset, but also to bypass the associated model storage cost that would normally be required.

Apart from the novel setting formulation, we showcase with experiments on large datasets that a vanilla OSOA algorithm can significantly save time versus training bespoke models and save space versus using one model for all targets.
Moreover, with the same adaptation step number or adaptation time, vanilla OSOA can exhibit better space efficiency, e.g., $47\%$ less space, than fine-tuning the pretrained model and saving the fine-tuned model.
We also show that further trade-off between time and space cost is available with varies strategies, e.g., multiple updates per batch and early stopping, and motivate more sophisticated algorithms with preliminary explorations.
Further, we use sample images from OSOA and baselines to justify adaptation based compression algorithms and to discuss societal considerations for DGMs based lossless compression.

The paper is organised as follows.
In Sec. \ref{sec:pre}, we review background knowledge regarding deep generative models for lossless compression.
In Sec. \ref{sec:osoa}, we formulate the setting of OSOA, discuss associated codec designing principles and propose a vanilla OSOA algorithm.
In Sec. \ref{sec:experiments}, we validate the setting by comparison with baselines, explore the strategies for further trade-off and discuss societal considerations.
In Sec. \ref{sec:related}, we discuss some related work. 
In Sec. \ref{sec:conclusion}, we summarise the paper, offer insights into our findings and point out promising future work directions.
\section{Preliminary}\label{sec:pre}
\textbf{Entropy Coding}
The core component underlying any lossless compression algorithm is the entropy coder, which converts between the data symbols and codewords.
Popular entropy coders include Huffman coding~\cite{huffman1952method}, arithmetic coding (AC)~\cite{witten1987arithmetic} and asymmetric numeral systems (ANS)~\cite{duda2013asymmetric}.
Huffman coding mainly involves binary trees operations, which is usually more time efficient than AC and ANS but can admit suboptimal codelength for probabilities not being power of $1/2$~\cite{mackay2003information}.
As a result, popular codecs adopted by DGM based algorithms are normally based on AC or ANS, e.g.,~\cite{mentzer2019practical,hoogeboom2019integer,mentzer2020learning}. 
The main difference between AC and ANS is that the former encodes and decodes in a first-in-first-out (FIFO) style while the latter performs in a first-in-last-out (FILO) style \cite{townsend2019practical}.
All entropy coders require the knowledge of the data symbol alphabet and the associated distribution, in order to conduct compression and decompression.
It is worth noting that the distribution used for entropy coding should be discrete and the average codelength is expected to be the entropy of the distribution used for coding.
Recall that for a discrete distribution with the probabilistic mass function $p(x)$, the (Shannon) entropy is defined as $\mathbf{E}_{x\sim p} [-\log p(x)]$.
Moreover, the optimal average codelength, given by the entropy of the ground truth distribution, is achieved only if the ground truth distribution is used for coding as per Shannon's source coding theorem \cite{mackay2003information}.
Specifically, compressing data from distribution $p(x)$ with distribution $q(x)$ will lead to the expected extra codelength at $\text{KL}(p(x)||q(x))$.

\textbf{Coding with Explicit Deep Generative Models} 
Since the entropy coders require probability values, implicit DGMs, e.g., generative adversarial networks (GANs)~\cite{goodfellow2014generative}, cannot be readily used for lossless compression.
For explicit DGMs, the learning principle is either maximum likelihood estimation  (MLE)~\cite{salimans2017pixelcnn++,kingma2018glow,ho2019flow++}:
\begin{equation}\label{eq:mle}
    \hat{\theta} = \arg\min_{\theta\in\Theta} \mathbf{E}_{x\sim p^*(x)} [-\log p_\theta (x)]
\end{equation}
or maximising the evidence lower bound (ELBO) of the logarithm evidence $\log p_\theta (x)$~\cite{kingma2013auto,townsend2019hilloc}:
\begin{equation}\label{eq:vi}
    \hat{\theta}, \hat{\phi} = {\arg\min}_{\theta\in\Theta, \phi\in\Phi} \mathbf{E}_{x\sim p^*(x)} [-\mathcal{L}(\theta, \phi, x)],
\end{equation}
where $\mathcal{L}(\theta, \phi, x)=\mathbf{E}_{q_\phi(z|x)}[-\log q_{\phi}(z|x) + \log p_\theta(x|z)p_\theta(z)]$.
Note that $x$ is a discrete random variable, so $p^*(x)$ refers to the probabilistic mass function of the ground truth data distribution.

When learning with MLE, one can either directly stipulate a discrete model or stipulate a continuous model with the dequantization technique~\cite{uria2013rnade, theis2015note}.
If one stipulates a discrete model, where $p_\theta(x)$ denotes the probabilistic mass function of the model, the objective function in Eq.~\ref{eq:mle} delegates the expected codelength when using $p_\theta(x)$ for entropy coding.
If one stipulates a continuous model, one normally dequantizes the discrete data $x$ with a continuous noise $u$ uniformly distributed on the unit hypercube, and gets the dequantized variable $y=x+u$ with distribution $p^*(y)$.
After fitting the probabilistic density function $p_\theta(y)$ of the model to the the probability density $p^*(y)$ of the dequantized distribution, one needs to discretise $p_\theta(y)$ and use the discretised distribution for entropy coding.
In this case, the MLE objective $\mathbf{E}_{y\sim p^*(y)} [-\log p_\theta (y)]$ delegates an upper bound of the expected codelength~\cite{theis2015note}.
For how to design sophisticated dequantization schemes or how to design discrete deep generative models, we refer the readers to \cite{ho2019flow++, hoogeboom2021learning, hoogeboom2019integer, berg2020idf++, ho2019compression, zhang2021ivpf}.

When learning with ELBO, one usually stipulates discrete models for $x$, i.e., $p_\theta(x|z)$ is a probabilistic mass function, and continuous models for $z$, i.e., $q_\phi(z|x)$ and $p_\theta(z)$ are probabilistic density functions.
Furthermore, one usually does not have direct access to the close form of $p_{\theta}(x)$ but the generative components $p_{\theta}(x|z)$ and $p_{\theta}(z)$.
To use $p_{\theta}(x|z)$ and $p_{\theta}(z)$ for efficient coding, a technique called bits-back \cite{wallace1990classification, hinton93keeping} is employed, which additionally involves the approximate posterior distribution $q_{\phi}(z|x)$ learned together with the generative components~\cite{townsend2019practical,townsend2019hilloc,kingma2019bit}.
In these cases, the discretisation precision of the latent variable will not effect the average codelength, delegated by the objective in Eq~\ref{eq:vi}, with consistent discretisation adopted for the latent distributions~\cite{townsend2019practical}.
It is worth noting that, since bits-back works more compatibly with ANS, bits-back ANS \cite{townsend2021lossless} is usually the first choice for models learned with the ELBO functional.

\section{One-shot online adaptation}\label{sec:osoa}
In this section, we formulate the setting into three stages, i.e., 1) Pretraining, 2) OSOA Encoding and 3) OSOA Decoding, and summarise the OSOA Encoding / Decoding stages in Alg.~\ref{alg:osoa_high}.
To unify the narrative of FIFO style and FILO style entropy coders, we aggregate the operations in Stage 2\&3 into two modules, namely the dynamical system module $\mathcal{D}$ and the codec module $\mathcal{C}$.
The deterministic dynamical system module $\mathcal{D}$ contains an optimiser, which can be a gradient based optimiser or a learning based meta optimiser, and associated parameters, to update a model $p_{\theta_{t-1}}(x)$ to $p_{\theta_{t}}(x)$ with a data batch $B_t$.
The codec module $\mathcal{C}$ contains an entropy coder, a codeword variable and methods to modify the codeword with the entropy coder, cf. following discussions and Alg. \ref{alg:encode}.

\begin{algorithm}[th!]
\small
\caption{One Shot Online Adaptation: Encoding and Decoding}
\begin{multicols}{2} 
\textbf{OSOA Encoding} 

\begin{algorithmic}[1]
\STATE {\bfseries Inbuilt attributes:} $p_{\theta_{0}}(x)$, $\mathcal{D}$, $\mathcal{C}$
\STATE {\bfseries Input:} data $\{x_i\}_{i=1}^N$
\STATE Form data batches $\{B_t\}_{t=1}^T$ from data $\{x_i\}_{i=1}^N$
\FOR{$t=1$ {\bfseries to} $T$}
\STATE $\mathcal{C}\texttt{.encode\_or\_cache}(p_{\theta_{t-1}}(x), B_t)$;
\STATE $p_{\theta_{t}}(x)=\mathcal{D}(p_{\theta_{t-1}}(x), B_t)$;
\ENDFOR
\STATE {\bfseries Output:} $c_f$ from $\mathcal{C}$.
\end{algorithmic}

\textbf{OSOA Decoding} 

\begin{algorithmic}[1]
\STATE {\bfseries Inbuilt attributes:} $p_{\theta_{0}}(x)$, $\mathcal{D}$, $\mathcal{C}$
\STATE {\bfseries Input:} code $c_f$
\STATE Initialise $\mathcal{C}$ with $c_f$
\FOR{$t=1$ {\bfseries to} $T$}
\STATE $B_t \leftarrow \mathcal{C}\texttt{.decode}(p_{\theta_{t-1}}(x))$;
\STATE $p_{\theta_{t}}(x)=\mathcal{D}(p_{\theta_{t-1}}(x), B_t)$;
\ENDFOR
\STATE {\bfseries Output:} data $\{x_i\}_{i=1}^N$ from $\{B_t\}_{t=1}^T$
\end{algorithmic}
\end{multicols}
\vspace{-8pt}
\label{alg:osoa_high}
\end{algorithm}

\textbf{Stage 1: Pretraining} 
A base model $p_{\theta_0}(x)$ is obtained by solving the problem defined in Eq. \ref{eq:mle} or Eq. \ref{eq:vi} on a pretraining dataset.

\textbf{Stage 2: OSOA Encoding}
A sequence of distributions $\{p_{\theta_t}(x)\}_{t=1}^{T}$ will be unrolled through the adaptation procedure by the deterministic dynamical system $\mathcal{D}$ with batches $\{B_t\}_{t=1}^T$.
Specifically, at each step $t$, one 1) inputs the model $p_{\theta_{t-1}}(x)$ and the data batch $B_t$ to the codec module and 2) performs the dynamical system $\mathcal{D}$ on $p_{\theta_{t-1}}(x)$ with $B_t$ and obtains the updated model $p_{\theta_{t}}(x)$.
In 1), the method $\mathcal{C}\texttt{.encode\_or\_cache}$ determines whether to directly encode the current batch (FIFO style) or save the needed information, i.e., the variable values and associated probabilistic mass function (pmf) values, for asynchronous encoding (FILO style), cf. Alg. \ref{alg:encode}.
At the end of OSOA Encoding, the final state of the codeword variable $c_{f}$ will be output from the codec module.

\textbf{Stage 3: OSOA Decoding}
One uses the codeword $c_f$ from the OSOA Encoding stage to initialise the codeword variable in $\mathcal{C}$ and starts the iterative decoding process as follows.
At each step $t$, one 1) uses the method $\mathcal{C}\texttt{.decode}$ to decode $B_t$ from the codeword variable with $p_{\theta_{t-1}}(x)$ and 2) updates the model with the decoded batch $B_t$ by the dynamical system $\mathcal{D}$.
As a result, the same sequence of distributions $\{p_{\theta_t}(x)\}_{t=1}^T$ will be unrolled and all the data batches $\{B_t\}_{t=1}^T$ will also be decoded.

\begin{figure}[ht!]
\centering
\setlength{\tabcolsep}{0pt}
\begin{tabular}{l}
\includegraphics[width=0.95\textwidth]{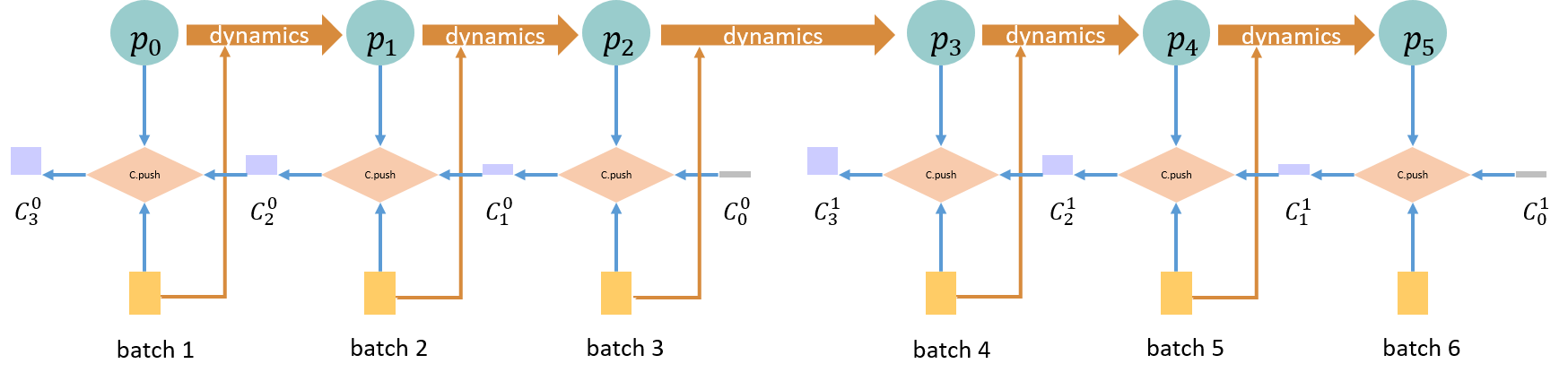}\\
\end{tabular}
\caption{An illustration of OSOA Encoding with FILO style entropy coders, where $m=3$ and $T=6$ (bb-ANS as an example). C.push denotes the encoding operation of the ANS codec. We refer readers to Appendix \ref{app:formulation} for complete demos of FIFO style and FILO style entropy coders.}
\label{fig:ans_encoding}
\end{figure}

\textbf{Methods in the codec module}
To recover the exact distribution sequence $\{p_{\theta_t}(x)\}_{t=1}^T$ in the decoding stage, one has to decode the data batches in the order of $\{B_1, B_2, ..., B_T\}$.
For DGMs based algorithms with an FIFO style coder, e.g., AC, one can consecutively encode the batch sequence $\{B_t\}_{t=1}^T$ in the original order and recover the correct order in the decoding stage.
However, for FILO style coders, e.g., ANS, one has to temporarily save associated pmf values for encoding purpose in cache and encode them with the entropy coder in the reverse order.
To reduce the size of cache memory, one can collect $m$ consecutive batches into one chunk and reversely compress the batches within the chunk, cf. Fig. \ref{fig:ans_encoding}.
In this way, one effectively compresses different chunks into different code files.
We refer readers to Appendix \ref{app:formulation} for complete demos of OSOA with FIFO and FILO style coders.
To distinguish the difference between the implementations of the \texttt{encode\_or\_cache} method in Alg.~\ref{alg:osoa_high} for coders of these two styles, we summarise the processes in Alg.~\ref{alg:encode}.
Note that \texttt{coder.encode} in Alg.~\ref{alg:encode} and $\mathcal{C}\texttt{.decode}$ in Alg.~\ref{alg:osoa_high} are respectively encoding and decoding functions already implemented with DGMs for lossless compression \cite{mentzer2019practical, townsend2019practical, townsend2019hilloc, kingma2019bit}.
Due to this flexibility, one can see that OSOA is generally applicable to DGMs based lossless compression algorithms.

\begin{algorithm}[th!]
\small
\caption{The \texttt{encode\_or\_cache} method in OSOA Encoding}
\begin{multicols}{2} 
\textbf{FIFO coders} 

\begin{algorithmic}[1]
\STATE {\bfseries Inbuilt attributes:} \texttt{coder}
\STATE {\bfseries Input:} $p_{\theta_{t-1}}(x), B_t$
\STATE $\texttt{coder.encode}(p_{\theta_{t-1}}(x), B_t)$;
\end{algorithmic}

\textbf{FILO coders} 

\begin{algorithmic}[1]
\STATE {\bfseries Inbuilt attributes:} \texttt{coder}, \texttt{cache}, $m$
\STATE {\bfseries Input:} $p_{\theta_{t-1}}(x), B_t$
\STATE \texttt{cache.append}($(p_{\theta_{t-1}}(x), B_t)$);
\IF{len(\texttt{cache}) $=m$ \textbf{or} \texttt{last\_batch}}
\FOR{$p(x), B$ in \texttt{reversed(cache)}}
\STATE $\texttt{coder.encode}(p(x), B)$;
\ENDFOR
\STATE \texttt{cache.empty()}
\ENDIF
\end{algorithmic}
\end{multicols}
\vspace{-8pt}
\label{alg:encode}
\end{algorithm}

\textbf{Example: Vanilla OSOA}
Following formulations of the proposed setting, we present the most straightforward strategy to conduct OSOA, referred to as vanilla OSOA.
Specifically, the base model $p_{\theta_0}(x)$ is pretrained on the same data modality, e.g., HiLLoC~\cite{townsend2019hilloc} pretrained on CIFAR10 for image compression;
and the deterministic dynamical system $\mathcal{D}$ is a gradient based optimiser, e.g., \texttt{AdaMax}~\cite{kingma2014adam}.
\textbf{Stage 1.} Find or train base model $p_{\theta_0}(x)$ for certain data modality.
\textbf{Stage 2.} Perform OSOA with base model from the last stage. The dataset to compress $\mathcal{S}$ should be first split to $T$ batches, i.e., $\mathcal{S}=\{B_t\}_{t=1}^T$. 
For $1\le t\le T$, we first use $p_{\theta_{t-1}}(x)$ to compute pmf values to encode the $t$-th batch. 
Then we use the $t$-th batch with the specific optimiser and certain learning rate to update and get the new model $p_{\theta_{t}}(x)$. 
\textbf{Stage 3.} Perform OSOA with base model from Stage 1. The data splitting strategy is the \textit{same} as Stage 2.
For $1\le t\le T$, we first use $p_{\theta_{t-1}}(x)$ to decode the $t$-th batch from the code.
Then we use the decoded batch with the \textit{same} $\mathcal{D}$ (e.g. optimiser, learning rate, etc.) as Stage 2 to update and get the new model $p_{\theta_{t}}(x)$.
Note the entropy coding part is omitted for narrative simplicity, and we refer readers to Appendix \ref{app:formulation} for more details.

\section{Experiments with vanilla OSOA}\label{sec:experiments}
\subsection{Vanilla OSOA versus baselines}\label{sec:baselines}
\textbf{Datasets} 
The datasets for base model pretraining are the renowned natural image datasets CIFAR10~\cite{krizhevsky2009learning} and ImageNet32~\cite{chrabaszcz2017downsampled}, including images of size $32\times 32$.
We obtain three target datasets randomly sampled from the large image dataset Yahoo Flickr Creative Commons 100 Million (YFCC100m)~\cite{thomee2016yfcc100m} to test the compression performance.
We name the three subsets from YFCC100m by SET32/64/128, where SET32 means the dataset of image height and width both being 32. 
To remove potential artifacts introduced by previous image compression, we first obtain SET256 by cropping the central $256\times 256$ part in $2^{17}$ RGB images with size larger than $256\times 256$ and downsample the obtained dataset into our targets, as conducted in \cite{mentzer2019practical}.
We use \texttt{Pillow} to downsample the SET256 into SET128, SET64 and SET32 with the high quality filter \texttt{Image.ANTIALIAS}.
The $2^{17}$ RGB images are selected only with their shapes and no other discretionary criteria.

\textbf{Experiments configurations}
Since an OSOA algorithm can be regarded as a general strategy to make a \textit{static} deep generative model \textit{dynamic}, it can be used for any DGMs based lossless compression frameworks.
We adopt two representative DGMs based compression frameworks, HiLLoC \cite{townsend2019hilloc} for hierarchical VAEs and IDF++ \cite{berg2020idf++} for normalizing flows, to validate the OSOA setting.
Specifically, for HiLLoC, we use the ResNet VAE (RVAE) model, which is proposed in \cite{kingma2016iaf} and adopted for compression in \cite{townsend2019hilloc}, with 24 layers and bi-directional connections for inference (a.k.a. top-down inference).
Moreover, we showcase for a proof-of-concept purpose the IAF RVAE \cite{kingma2016iaf}, with the same configuration as the RVAE model except for additional IAF layers in the inference model, making the approximate posterior more expressive.
For IDF++, we use the same architecture proposed in \cite{berg2020idf++}, with 24 stacked flow layer blocks, where each block contains a permutation layer and a coupling layer. 
For each coupling layer in the flow model, a 12-layer Densenet \cite{huang2017densely} with 512 feature maps is used.
For narrative simplicity, we use HiLLoC and IDF++ to refer to both the frameworks and the associated models.
We use an Nvidia V100 32GB GPU for HiLLoC (and IAF RVAE) and an Nvidia V100 16 GB GPU for IDF++.
Capped by the GPU memory limit, the batch size should be reduced as the image size grows.
We quadruple the batch size as the image size decreases, i.e., batch size 256/64/16 in HiLLoC and batch size 48/12/3 in IDF++, for SET32/64/128 respectively. 

\begin{table}[th!]
\caption{Theoretical bits per dimension (bpd) values of HiLLoC, IAF RVAE and IDF++ models training from scratch (ReTrain) on the three target datasets SET32/64/128.}
\label{tab:retrain}
\begin{center}
\begin{tabular}{cccccccccc}
\toprule
\multicolumn{3}{c}{HiLLoC} & \multicolumn{3}{c}{IAF RVAE} & \multicolumn{3}{c}{IDF++} \\
\cmidrule(lr){1-3}\cmidrule(lr){4-6}\cmidrule(lr){7-9}
SET32 & SET64 & SET128 & SET32 & SET64 & SET128 & SET32  & SET64 & SET128 \\
\midrule
2.668 & 2.504 & 2.521  & 2.608 & 2.292 & 2.392  & 2.361  & 2.443 & 2.422 \\
\bottomrule
\end{tabular}
\end{center}
\end{table}

\begin{table}[t]
\caption{Theoretical bpd values between vanilla OSOA and baselines (PreTrain, FineTune v1, v2 and v3). The bpd values of baselines less than those of vanilla OSOA are shown in green, e.g., $\FG{0.132}=3.505-3.373$. The effective bpd values by saving the models with FineTune baselines, defined as $\#\texttt{trainable\_parameters}\times\texttt{bits/parameter}\times 1/\#\texttt{dataset\_total\_dims}$, are shown in red. HiLLoC CIFAR10 denotes the CIFAR10 pretrained HiLLoC model.}
\label{tab:main_baseline}
\setlength{\tabcolsep}{5pt}
\begin{center}
\begin{tabular}{lccllccc}
\toprule
        & SET & PreTrain & OSOA  & FineTune v1        & FineTune v2        & FineTune v3       & Model \\
\midrule
        & 32  & 3.830    & 3.505 & 3.373 \FG{(0.132)} & 3.322 \FG{(0.183)} & 3.215 \FG{(0.290)} & \RD{3.258}\\
HiLLoC  & 64  & 3.561    & 3.090 & 2.954 \FG{(0.136)} & 2.907 \FG{(0.183)} & 2.800 \FG{(0.290)} & \RD{0.815}\\
CIFAR10 & 128 & 3.268    & 2.682 & 2.564 \FG{(0.118)} & 2.518 \FG{(0.164)} & 2.412 \FG{(0.270)} & \RD{0.204}\\
\midrule
        & 32  & 3.739    & 3.537 & 3.398 \FG{(0.139)} & 3.329 \FG{(0.208)} & 3.196 \FG{(0.341)} & \RD{3.258}\\
HiLLoC  & 64  & 3.430    & 3.109 & 2.950 \FG{(0.159)} & 2.893 \FG{(0.216)} & 2.769 \FG{(0.340)} & \RD{0.815}\\
ImgNet32& 128 & 3.104    & 2.682 & 2.534 \FG{(0.148)} & 2.482 \FG{(0.200)} & 2.362 \FG{(0.320)} & \RD{0.204}\\
\midrule
IAF     & 32  & 3.930    & 3.378 & 3.245 \FG{(0.133)} & 3.185 \FG{(0.193)} & 3.059 \FG{(0.319)} & \RD{3.963}\\
RVAE    & 64  & 4.078    & 3.028 & 2.816 \FG{(0.212)} & 2.751 \FG{(0.277)} & 2.640 \FG{(0.388)} & \RD{0.991}\\
CIFAR10 & 128 & 4.149    & 2.596 & 2.403 \FG{(0.193)} & 2.354 \FG{(0.242)} & 2.250 \FG{(0.346)} & \RD{0.248}\\
\midrule
IAF     & 32  & 3.551    & 3.351 & 3.208 \FG{(0.143)} & 3.146 \FG{(0.205)} & 3.012 \FG{(0.339)} & \RD{3.963}\\
RVAE    & 64  & 3.325    & 2.932 & 2.769 \FG{(0.163)} & 2.707 \FG{(0.225)} & 2.582 \FG{(0.350)} & \RD{0.991}\\
ImgNet32& 128 & 3.037    & 2.502 & 2.354 \FG{(0.148)} & 2.300 \FG{(0.202)} & 2.183 \FG{(0.319)} & \RD{0.248}\\
\midrule
        & 32  & 3.612    & 3.225 & 3.089 \FG{(0.136)} & 3.045 \FG{(0.180)} & 2.835 \FG{(0.390)} & \RD{4.517} \\
IDF++   & 64  & 3.471    & 2.873 & 2.725 \FG{(0.148)} & 2.687 \FG{(0.186)} & 2.481 \FG{(0.392)}                   & \RD{1.166} \\
CIFAR10 & 128 & 3.441    & 2.554 & 2.374 \FG{(0.180)} & 2.335 \FG{(0.219)} & -                  & \RD{0.292} \\
\midrule
        & 32  & 3.662    & 3.338 & 3.160 \FG{(0.178)} & 3.119 \FG{(0.219)} & 2.921 \FG{(0.417)} & \RD{4.517}\\
IDF++   & 64  & 3.439    & 2.949 & 2.768 \FG{(0.181)} & 2.729 \FG{(0.220)} & 2.528 \FG{(0.421)}                   & \RD{1.166}\\
ImgNet32& 128 & 3.401 & 2.623  & 2.448 \FG{(0.175)} & 2.404 \FG{(0.219)} & -  & \RD{0.292}\\
\bottomrule
\end{tabular}
\end{center}
\end{table}

\textbf{Baselines} 
We adopt three baselines, referred to as ReTrain, PreTrain and FineTune.
For ReTrain, we train a separate model for each target dataset from scratch and use the final model to compress the target datasets.
For PreTrain, we directly use the pretrained models to compress the target datasets.
For FineTune, we fine tune the pretrained model on the target datasets and use the final model to compress the target datasets.
Depending on how to set the number of fine-tuning epoches, we have three versions of FineTune.
For FineTune v1, we fine tune the pretrained model for 2 epochs, as the whole OSOA Encoding \& Decoding procedures involve 2 epochs of adaptations in total.
For FineTune v2, we fine tune the pretrained model for 4 epochs for HiLLoC (and IAF RVAE) and 3 epochs for IDF++ to make the model updating time comparable with OSOA.
This is because the dynamical system module in vanilla OSOA needs strict determinism \cite{dldet} and it could cause extra time cost in current deep learning frameworks.
The time ratio we measured with/without the determinism is 1.98 (HiLLoC) in \texttt{TensorFlow 1.14} \cite{abadi2016tensorflow} with \texttt{tensorflow-determinism 0.3.0} \cite{tfdet} and 1.34 (IDF++) in \texttt{Pytorch 1.6} \cite{torchdet}.
For FineTune v3, we fine tune the pretrained model for 20 epochs as an indicator of overfitting the pretrained model on target datasets.
To take the additional model storage in FineTune baselines into consideration, we convert the model (in \texttt{float32}) storage size into effective bpd values and show them in the Model column in Table~\ref{tab:main_baseline}.
Note that the bpd values shown under OSOA and baselines are of the codewords only, and the associated model bpd in the Model column need to be added to ReTrain and FineTune baselines for comparison of total storage cost.
As bpd is the average number of bits needed to compress each dimension of the data, a lower value indicates a higher compression ratio.
The time costs of OSOA and baselines are caused by model training, network inference (pmf evaluation) and entropy coding.
We predominately use the model training time for comparison, as it dominates the time cost and the model and the entropy coder are controlled factors for comparisons.

Recent work \cite{nalisnick2018deep, choi2018waic, nalisnick2019detecting} show that DGMs can assign high density values on data points not in the training datasets. 
And pretrained models can provide reasonable compression performance on unseen datasets \cite{townsend2019hilloc, zhang2021ivpf}.
While PreTrain in Table \ref{tab:main_baseline} further confirms this point, the extra cost comparing to best possible ones can be significant, due to the distribution discrepancy between the training data and the target data.
Nevertheless, PreTrain has the best time efficiency, since no extra training is needed.

\begin{figure}[ht!]
\centering
\setlength{\tabcolsep}{3pt}
\begin{tabular}{ccc}
\includegraphics[width=0.31\textwidth]{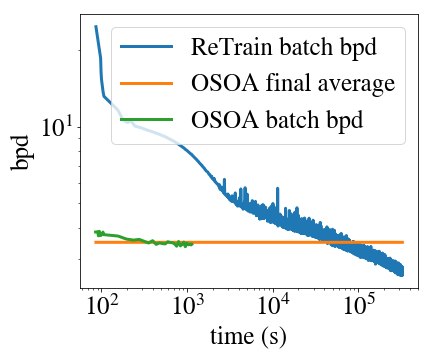} &
\includegraphics[width=0.31\textwidth]{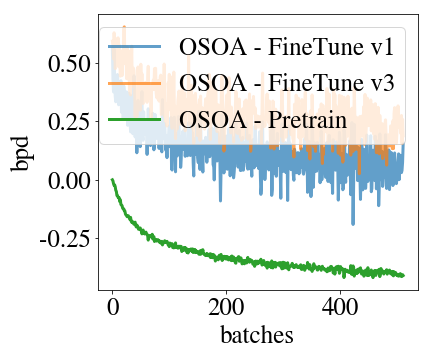} &
\includegraphics[width=0.31\textwidth]{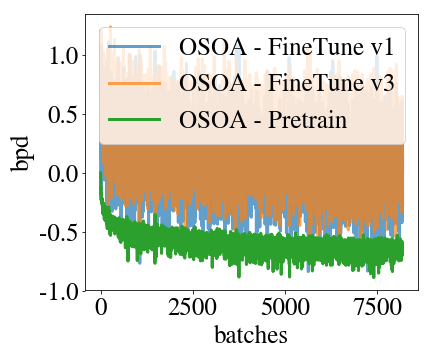}\\
\end{tabular}
\caption{Left: Theoretical bpd values of each batch during training HiLLoC from scratch on SET32 (ReTrain HiLLoC) versus the batch bpd of OSOA, showing the significant time cost of training from scratch. OSOA final average is the bpd value after compressing all the batches with OSOA. Middle and Right: the differences between the theoretical bpd values of OSOA and the theoretical bpd values of the baselines of each batch in SET32 (Middle) and SET128 (Right) with the CIFAR10 pretrained HiLLoC. Negative values indicate the advantages of coding with OSOA instead of the corresponding baseline. The disadvantage of OSOA (versus FineTune) tends to decrease and the advantage of OSOA (versus PreTrain) tends to increase as the online adaptation being conducted.}
\label{fig:bpd_hist}
\end{figure}

Comparing to FineTune, OSOA does not require additional model storage for decompression.
Fixing the batch order, the unrolled models in OSOA are the models in the first fine-tuning epoch, which tend to fit the data worse than the final one in the whole fine-tuning process (not necessarily true, cf. negative difference values in Fig \ref{fig:bpd_hist} Right).
Nevertheless, by comparing the decreased code space shown in green and the increased model space shown in red for FineTune v1 or FineTune v2 in Table \ref{tab:main_baseline}, one can see that with the same optimisation budget or time budget, OSOA exhibits better total space efficiency than FineTune.
For instance, with the CIFAR10 pretrained HiLLoC on SET32, OSOA takes $46.73\%$ less space than FineTune v2, with the same model training cost.
{To better understand the comparison between OSOA and FineTune v2 on SET128, we take the CIFAR10 pretrained HiLLoC and fix the order of images.
If we only compress the first $2^{16}$, $2^{15}$ and $2^{14}$ images in SET128, the FineTune v2 admits bpd, in the code bpd + model bpd format, 2.564 + 0.408, 2.611 + 0.816 and 2.658 + 1.632. 
However, OSOA admits bpd, in the code bpd format since no model storage required, 2.734, 2.788 and 2.842, which are 0.238, 0.639 and 1.448 less than FineTune v2 respectively. 
Therefore, OSOA remains a clear net advantage than FineTune with the same time budget up to a reasonable large sample size even on large size images.}
From the comparison between OSOA and FineTune v3, one can see that the space advantage is still valid even if we conduct the FineTune for much longer time for SET32/64.
FineTune v3 performs better on space efficiency on SET128, but it takes more than 4 times as much time as OSOA.
Recent work on data lossy compression \cite{van2021overfitting} introduces a model update coding term to the objective function and conducts lossless compression on the model update.
However, this introduces an extra layer of complexity and time cost to determine suitable prior distributions for network parameters and to balance the data compression and model compression.
Since OSOA does not require extra model saving, it exhibits huge advantage over the baselines on large-scale models or data of limited size.
Note that OSOA achieves the optimal space \& time cost balance with conducting model updates during data decoding.
Therefore, it will be more suitable to scenarios of data backup or data cold storage than scenarios with real time data usage, e.g., video streaming.

At each OSOA step, the potential improvement of the model from the current batch can only benefit the compression of following batches.
Moreover, the batch gradient can have a large variance and the stochastic gradient methods, e.g. SGD, Adam, etc., are not strictly descent algorithms.
To investigate the online compression performance along the OSOA path, we showcase the results at the batch level in OSOA Encoding in Fig. \ref{fig:bpd_hist} Middle and Right, with the CIFAR10 pretrained HiLLoC on SET32 and SET128.
To make the bpd values comparable, we use the same batch order for OSOA, FineTune and PreTrain and showcase the differences between the bpd values of compressing each batch with OSOA and the baselines.
Here we abuse the notation and use OSOA/FineTune to denote the vector of bpd values of each batch in OSOA/FineTune, with above fixed batch order.
Then the 100th entry in the difference vector (OSOA - FineTune v1) is the extra cost of compressing the 100th batch with OSOA at the 100th OSOA Encoding step instead of FineTune v1 (with the fine-tuned model).
Note that a negative difference value indicates that OSOA performs better than the corresponding baseline for that batch.
From Fig. \ref{fig:bpd_hist} Middle and Right, we can see that the advantage of compression with OSOA instead of PreTrain enlarges and the disadvantage (w.r.t. code space) of compression with OSOA instead of FineTune decreases, which validates the benefits brought by adaptation with previous batches to following batches in OSOA.

Note that the theoretical bpd values reported in this paper are evaluated based on the discretised distributions of models for coding.
Moreover, for HiLLoC, although the free-bits technique is used for the ResNet VAE model training, theoretical bpd evaluation does not involve free-bits.
We showcase the real bpd values of HiLLoC and IDF++ on SET32 in Appendix \ref{app:experiments}.

\subsection{Vanilla OSOA variants to prioritise space or time efficiency}\label{sec:variants}

\textbf{OSOA with multiple updates per batch}
Vanilla OSOA conducts one model update step per data batch with gradient based optimisers.
However, it might not be sufficient for OSOA to exploit the adaptation potential from the base model, which is especially true for scenarios with small number of data batches.
For instance, by comparing Fig. \ref{fig:bpd_hist} Middle and Right, we can see that the performance improvement curves exhibit slower overall convergences on SET32 than SET128.
To investigate the potential of OSOA in such scenarios, we follow the setting in Sec.~\ref{sec:baselines} and conduct vanilla OSOA with multiple updates per batch.
Note that each data batch is only used in one corresponding OSOA step and the multiple updates are conducted in that OSOA step, which will not change the one-shot nature and is different from multiple-epoch adaptation.
In this section, we use SET32 with CIFAR10 pretrained HiLLoC and IDF++ and fix the batch size at 256 and 48 respectively.
From Table~\ref{tab:many_updates}, we can see that conducting multiple updates per batch can improve the performance for both HiLLoC and IDF++, and it can even further improve the bpd by around $0.4$ for HiLLoC.
It is worth noting that in contrast to that overfitting does not happend for HiLLoC at 500 steps per batch, overfitting happens at 100 steps per batch for IDF++.
Moreover, conducting multiple updates per batch scales the time cost linearly and it provides a modification to trade off time efficiency for space efficiency.

\begin{table}[th!]
\caption{The theoretical bpd of HiLLoC and IDF++ with OSOA of multiple optimisation updates per batch on test dataset SET32.}
\label{tab:many_updates}
\begin{center}
\begin{tabular}{lccccccc}
\toprule
updates/batch    & 1     & 3     & 5     & 7     & 10    & 100   & 500   \\
\midrule
HiLLoC (CIFAR10) & 3.505 & 3.420 & 3.382 & 3.357 & 3.332 & 3.184 & 3.109  \\
IDF++ (CIFAR10)  & 3.225 & 3.137 & 3.096 & 3.080 & 3.046 & 3.105 & - \\
\bottomrule
\end{tabular}
\end{center}
\end{table}

\textbf{OSOA with early stopping}
In vanilla OSOA, the number of adaptation steps is determined by the data size and the batch size.
For very large datasets, additional bpd improvements from the latter OSOA steps may not be so significant, cf. Fig~\ref{fig:bpd_hist} Right.
Due to the strict reproducibility of OSOA, if one sets a stopping criterion and uses the model at the stopping time for remaining data batches, the stopping criterion will be triggered at exactly the same step in OSOA Encoding and OSOA Decoding.
Moreover, the time for model update, i.e., back propagation in the vanilla OSOA, will be saved for remaining batches.
For instance, in the example of CIFAR10 pretrained HiLLoC for SET128, there are $8192$ batches in total.
If one stops the model adaptation at the $500$th, $1000$th, $2500$th or $5000$th batch, the final OSOA bpd will be $2.820$, $2.774$, $2.717$ or $2.690$ respectively, where one can accelerate the model adaptation efficiency by 8 times with the bpd cost less than $0.1$.

\begin{figure*}[ht!]
\centering
\setlength{\tabcolsep}{3pt}
\begin{tabular}{ccc}
\includegraphics[width=0.31\textwidth]{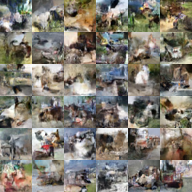} &
\includegraphics[width=0.31\textwidth]{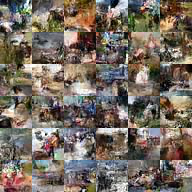} &
\includegraphics[width=0.31\textwidth]{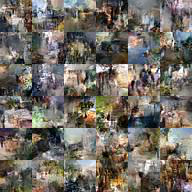}\\
(a) PreTrain & (b) FineTune v3 & (c) ReTrain\\
\includegraphics[width=0.31\textwidth]{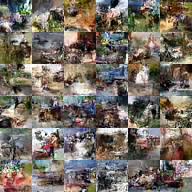} &
\includegraphics[width=0.31\textwidth]{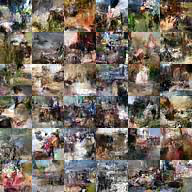} &
\includegraphics[width=0.31\textwidth]{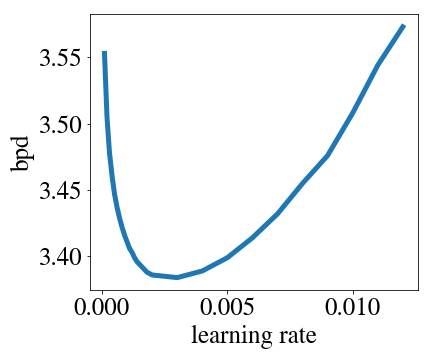}\\
(d) OSOA 1 update/batch & (e) OSOA 100 updates/batch & (f) OSOA bpd vs. lr
\end{tabular}
\caption{36 images sampled from (a) PreTrain: CIFAR10 pretrained HiLLoC, (b) FineTune v3: fine-tuning CIFAR10 pretrained HiLLoC for 20 epochs on SET32, (c) ReTrain: HiLLoC trained on SET32 from scratch, (d) and (e) OSOA 1 and 100 updates/batch: the final checkpoint of vanilla OSOA with 1 and 100 update steps per batch from CIFAR10 pretrained HiLLoC, respectively. (f) The theoretical bpd values of vanilla OSOA with different learning rate values, with HiLLoC (CIFAR10).}
\label{fig:imgs}
\end{figure*}

\subsection{Discussion on DGMs for data lossless compression}\label{sec:imgs}
Probabilistic modelling with DGMs for data generation encourages precise modelling of the target dataset for high fidelity generation.
Different from generation purpose modelling, lossless compression purpose modelling only calls for high density values.
The phenomena, DGMs assigning high density values for out-of-distribution (OOD) data \cite{nalisnick2018deep, choi2018waic, nalisnick2019detecting} and plausible performance of pretrained models for OOD datasets lossless compression \cite{townsend2019hilloc, zhang2021ivpf}, indicate that data generation purpose modelling and data compression purpose modelling are not strictly equivalent.
And we refer the readers to \cite{theis2015note} for an intuitive discussion on sample quality and log-likelihood values.
It is discussed in \cite{nalisnick2018deep} that DGMs weigh low-level statistics more than high-level semantics, and we show in above sections and Fig.~\ref{fig:imgs} such inductive bias can be leveraged for lossless compression purpose.
Derived from PreTrain, OSOA and FineTune admit samples with consistent semantics as PreTrain, fixing the random seed, cf. Fig.~\ref{fig:imgs} (a), (b), (d) and (e).
At the local feature level, e.g., sharpness and contrast, samples from the final models of OSOA and FineTune are more similar to samples from ReTrain, explaining improvements of OSOA and FineTune from PreTrain.
While such local feature transfer largely originates from the small learning rate used in OSOA and FineTune, learning rate of too small or too large values will impede the performance, cf. Fig.~\ref{fig:imgs} (f).
Note that (b), (d) and (e) in Fig.~\ref{fig:imgs} are hardly distinguishable, but they do differ in numerical values.

Another motivation of presenting samples in Fig.~\ref{fig:imgs} is to discuss societal considerations to be addressed for future real applications of DGMs based lossless compression algorithms.
Since trained DGMs model the distribution of the training set, information of the training data may be leaked by samples from the trained models, e.g., the third row and fifth column in Fig.~\ref{fig:imgs} (c) showing two people.
Therefore, the specially trained models should be protected from privacy violation.
From this perspective, pretrained models with as few adaptation steps as possible tend to have more privacy protection for service users.
Nevertheless, information of the pretraining dataset can be retained in adapted models and influence the compression performance of target datasets.
As a result, service providers should avoid using sensitive information in the pretraining dataset and avoid discrimination (different compression performances for data of different groups) caused by the pretraining dataset.
As shown in this work, training on low resolution and non-sensitive images can offer good performance on images of different resolutions and semantics.
Accordingly, it is promising to build societally friendly and effective DGMs base lossless compression algorithms with pretraining + adaptation style frameworks, e.g., OSOA, and careful incorporation of pretraining data.

\section{Related work}\label{sec:related}
OSOA de facto defines a paradigm to compress the distribution (model) sequence.
The high level principle of it can trace back to a simple mathematical fact: a sequence can be uniquely determined by an initial point and a deterministic transition dynamics.
In this section, we provide a non-exhaustive list of how this high level principle can be instantiated in compression related literature.

\cite{schmidhuber1996sequential, schmidhuber1995predictive} showcased the potential of neural networks as static context models for text compression and discussed the possibility of adaptive modelling with neural networks.
\cite{mahoney2000fast} and DeepZip \cite{tatwawadi2018deepzip, goyal2018deepzip} then instantiated the above principle for neural network based adaptive modelling with AC.
This line of work focuses on the sequential modelling of entries in one data point and thus is intra-data adaptive modelling; we focus on DGMs of the data distribution and thus is inter-data adaptive modelling.
Moreover, our work additionally bridges the initial probabilistic model with the adaptation process which enables the adaptation of a pretrained model and the benefit from pretrained models of datasets sharing common structures.
Another line of work is sequential data lossy compression like videos~\cite{van2021overfitting,habibian2019video}. 
They introduce encoder-decoder networks for compression, where the encoder network is adapted to sequential data during compression and the decoder network is used for decompression.
However, the setting is quite different from OSOA. 
First, they are the lossy compression method and could only use VAEs for compression, while ours mainly deals with lossless ones and works on any explicit generative model. 
Second, they mainly deal with sequential data, while ours can compress any data batches, including the sequential ones.

The model storage cost for deep learning based data lossless compression is also considered in \cite{blier2018description}. 
The work~\cite{blier2018description} formulates the problem in a supervised learning paradigm, where both the sender and receiver need the base model and an auxiliary dataset for the label transmission, whereas our work formulates in the unsupervised learning paradigm and only the base model is needed.
Moreover, in order for the auxiliary dataset needed in \cite{blier2018description} to effectively reduce the bitrate of the label (data to transfer), the label data needs high mutual information with the auxiliary dataset, which can form a stronger constraint.

The same principle also applies to pseudo-random generators, where a unique sample can be identified by the index of the sample generated from a shared pseudo-random source.
\cite{havasi2018minimal} leveraged this fact and proposed an autoregressive style algorithm to compress the network parameter partition by partition with the sample indices of each partition and a predefined parameter prior.
In the setting of lossy compression of Bayesian neural networks, \cite{havasi2018minimal} chooses one sample from the learned approximate posterior distribution to compress and thus deviates from the setting of data lossless compression where the data is given instead of chosen.

Prior to the developments in the deep learning literature, this high level principle has also been investigated for data compression.
Adaptive dictionary methods, e.g., LZ77~\cite{ziv1977universal}, dynamically maintain a dictionary of substrings to gradually recover the optimal dictionary containing the repeated patterns in the sequence.
Apart from dictionaries, the probability simplex can also be dynamically maintained to gradually converge to the ground truth probabilities or to capture the distribution shifts, e.g., context-adaptive binary arithmetic coding (CABAC)~\cite{pennebaker1988overview, marpe2003context}.
Furthermore, for context mixture modelling methods, the mixture weights can be adapted during compression to favour the context models that make more accurate predictions, e.g., PAQ~\cite{mahoney2005adaptive}.
Due to the functional form of deep generative models, conducting OSOA of DGMs for lossless compression uniquely leverages the inductive bias of DGMs and could admit more flexible compression purpose modelling.

\section{Conclusion and future work}\label{sec:conclusion}
In this work, we proposed the novel setting of One-Shot Online Adaptation (OSOA) of deep generative models for lossless compression and discussed codec designing principles for codecs of different styles.
We showcased with a vanilla OSOA algorithm that OSOA can reduce the extensive model training time and model storage space for new datasets and achieve optimal trade-off between time and space cost versus alternative solutions.
Our experimental exploration indicated that more sophisticated OSOA algorithms are promising and desirable.
Finally, we discussed societal considerations for deep generative models based lossless compression algorithms.

The future work can be pursued in three folds.
First, since the quality of the base model is very important for reducing the OSOA cost, it is highly desirable to study how to use deep generative models to capture the most of the common information shared among a general class of different datasets.
Second, as shown in Sec. \ref{sec:variants}, there is still room to improve the space and time efficiency within one epoch of adaptation, it is thus of interest to study how more advanced dynamical systems, e.g., meta-learning methods or reinforcement learning methods, can further exploit the full potential of OSOA. 
Third, following the discussion in Sec. \ref{sec:imgs}, it is interesting to investigate local feature transfer algorithms with pretrained models for fast lossless compression on unseen datasets.

\begin{ack}
The authors thank the anonymous reviewers for their review and suggestions, which improved the quality of this manuscript.
The authors thank Ning Kang and Mingtian Zhang for fruitful discussions.
\end{ack}

\newpage
\bibliography{osoa}

\begin{thebibliography}{10}

\bibitem{dldet}
Determinism in deep learning.
\newblock \url{https://developer.nvidia.com/gtc/2019/video/s9911}.

\bibitem{torchdet}
Pytorch reproducibility.
\newblock \url{https://pytorch.org/docs/stable/notes/randomness.html}.

\bibitem{tfdet}
Tensorflow determinism.
\newblock \url{https://github.com/NVIDIA/framework-determinism}.

\bibitem{abadi2016tensorflow}
Mart\'{\i}n Abadi, Paul Barham, Jianmin Chen, Zhifeng Chen, Andy Davis, Jeffrey
  Dean, Matthieu Devin, Sanjay Ghemawat, Geoffrey Irving, Michael Isard,
  Manjunath Kudlur, Josh Levenberg, Rajat Monga, Sherry Moore, Derek~G. Murray,
  Benoit Steiner, Paul Tucker, Vijay Vasudevan, Pete Warden, Martin Wicke, Yuan
  Yu, and Xiaoqiang Zheng.
\newblock Tensorflow: A system for large-scale machine learning.
\newblock In {\em Proceedings of the 12th USENIX Conference on Operating
  Systems Design and Implementation}, OSDI'16, page 265–283, USA, 2016.
  USENIX Association.

\bibitem{blier2018description}
L\'{e}onard Blier and Yann Ollivier.
\newblock The description length of deep learning models.
\newblock In {\em Advances in Neural Information Processing Systems},
  volume~31, 2018.

\bibitem{choi2018waic}
Hyunsun Choi, Eric Jang, and Alexander~A Alemi.
\newblock Waic, but why? generative ensembles for robust anomaly detection.
\newblock {\em arXiv preprint arXiv:1810.01392}, 2018.

\bibitem{chrabaszcz2017downsampled}
Patryk Chrabaszcz, Ilya Loshchilov, and Frank Hutter.
\newblock A downsampled variant of imagenet as an alternative to the cifar
  datasets.
\newblock {\em arXiv preprint arXiv:1707.08819}, 2017.

\bibitem{collet2016smaller}
Yann Collet and Chip Turner.
\newblock Smaller and faster data compression with zstandard.
\newblock {\em Facebook Code [online]}, 2016.

\bibitem{duda2013asymmetric}
Jarek Duda.
\newblock Asymmetric numeral systems: entropy coding combining speed of huffman
  coding with compression rate of arithmetic coding.
\newblock {\em arXiv preprint arXiv:1311.2540}, 2013.

\bibitem{gage1994new}
Philip Gage.
\newblock A new algorithm for data compression.
\newblock {\em C Users Journal}, 12(2):23--38, 1994.

\bibitem{goodfellow2014generative}
Ian Goodfellow, Jean Pouget-Abadie, Mehdi Mirza, Bing Xu, David Warde-Farley,
  Sherjil Ozair, Aaron Courville, and Yoshua Bengio.
\newblock Generative adversarial nets.
\newblock {\em Advances in Neural Information Processing Systems}, 27, 2014.

\bibitem{goyal2018deepzip}
Mohit Goyal, Kedar Tatwawadi, Shubham Chandak, and Idoia Ochoa.
\newblock Deepzip: Lossless data compression using recurrent neural networks.
\newblock In {\em Data Compression Conference, {DCC} 2019, Snowbird, UT, USA,
  March 26-29, 2019}, page 575. {IEEE}, 2019.

\bibitem{habibian2019video}
Amirhossein Habibian, Ties~van Rozendaal, Jakub~M Tomczak, and Taco~S Cohen.
\newblock Video compression with rate-distortion autoencoders.
\newblock In {\em Proceedings of the IEEE/CVF International Conference on
  Computer Vision}, pages 7033--7042, 2019.

\bibitem{havasi2018minimal}
Marton Havasi, Robert Peharz, and José~Miguel Hernández-Lobato.
\newblock Minimal random code learning: Getting bits back from compressed model
  parameters.
\newblock In {\em International Conference on Learning Representations}, 2019.

\bibitem{hinton93keeping}
GE~Hinton and Drew van Camp.
\newblock Keeping neural networks simple by minimising the description length
  of weights. 1993.
\newblock In {\em Proceedings of COLT-93}, pages 5--13.

\bibitem{ho2019flow++}
Jonathan Ho, Xi~Chen, Aravind Srinivas, Yan Duan, and Pieter Abbeel.
\newblock Flow++: Improving flow-based generative models with variational
  dequantization and architecture design.
\newblock In {\em International Conference on Machine Learning}, pages
  2722--2730. PMLR, 2019.

\bibitem{ho2019compression}
Jonathan Ho, Evan Lohn, and Pieter Abbeel.
\newblock Compression with flows via local bits-back coding.
\newblock In {\em Advances in Neural Information Processing Systems}, pages
  3879--3888, 2019.

\bibitem{hoogeboom2021learning}
Emiel Hoogeboom, Taco Cohen, and Jakub~Mikolaj Tomczak.
\newblock Learning discrete distributions by dequantization.
\newblock In {\em Third Symposium on Advances in Approximate Bayesian
  Inference}, 2021.

\bibitem{hoogeboom2019integer}
Emiel Hoogeboom, Jorn Peters, Rianne van~den Berg, and Max Welling.
\newblock Integer discrete flows and lossless compression.
\newblock In {\em Advances in Neural Information Processing Systems}, pages
  12134--12144, 2019.

\bibitem{huang2017densely}
Gao Huang, Zhuang Liu, Laurens Van Der~Maaten, and Kilian~Q Weinberger.
\newblock Densely connected convolutional networks.
\newblock In {\em Proceedings of the IEEE Conference on Computer Vision and
  Pattern Recognition}, pages 4700--4708, 2017.

\bibitem{huffman1952method}
David~A Huffman.
\newblock A method for the construction of minimum-redundancy codes.
\newblock {\em Proceedings of the IRE}, 40(9):1098--1101, 1952.

\bibitem{kingma2014adam}
Diederik~P. Kingma and Jimmy Ba.
\newblock Adam: {A} method for stochastic optimization.
\newblock In {\em 3rd International Conference on Learning Representations,
  {ICLR} 2015}, 2015.

\bibitem{kingma2013auto}
Diederik~P. Kingma and Max Welling.
\newblock Auto-encoding variational bayes.
\newblock In {\em 2nd International Conference on Learning Representations,
  {ICLR} 2014}.

\bibitem{kingma2018glow}
Durk~P Kingma and Prafulla Dhariwal.
\newblock Glow: Generative flow with invertible 1x1 convolutions.
\newblock In {\em Advances in Neural Information Processing Systems}, pages
  10215--10224, 2018.

\bibitem{kingma2016iaf}
Durk~P Kingma, Tim Salimans, Rafal Jozefowicz, Xi~Chen, Ilya Sutskever, and Max
  Welling.
\newblock Improved variational inference with inverse autoregressive flow.
\newblock In D.~Lee, M.~Sugiyama, U.~Luxburg, I.~Guyon, and R.~Garnett,
  editors, {\em Advances in Neural Information Processing Systems}, volume~29,
  pages 4743--4751. Curran Associates, Inc., 2016.

\bibitem{kingma2019bit}
Friso Kingma, Pieter Abbeel, and Jonathan Ho.
\newblock Bit-swap: Recursive bits-back coding for lossless compression with
  hierarchical latent variables.
\newblock In {\em International Conference on Machine Learning}, pages
  3408--3417. PMLR, 2019.

\bibitem{knuth1985dynamic}
Donald~E Knuth.
\newblock Dynamic huffman coding.
\newblock {\em Journal of algorithms}, 6(2):163--180, 1985.

\bibitem{krizhevsky2009learning}
Alex Krizhevsky, Geoffrey Hinton, et~al.
\newblock Learning multiple layers of features from tiny images.
\newblock 2009.

\bibitem{kumar2019videoflow}
Manoj Kumar, Mohammad Babaeizadeh, Dumitru Erhan, Chelsea Finn, Sergey Levine,
  Laurent Dinh, and Durk Kingma.
\newblock Videoflow: A conditional flow-based model for stochastic video
  generation.
\newblock In {\em International Conference on Learning Representations}, 2020.

\bibitem{mackay2003information}
David~JC MacKay.
\newblock {\em Information theory, inference and learning algorithms}.
\newblock Cambridge university press, 2003.

\bibitem{mahoney2000fast}
Matthew~V Mahoney.
\newblock Fast text compression with neural networks.
\newblock In {\em FLAIRS conference}, pages 230--234, 2000.

\bibitem{mahoney2005adaptive}
Matthew~V Mahoney.
\newblock Adaptive weighing of context models for lossless data compression.
\newblock Technical report, 2005.

\bibitem{marpe2003context}
Detlev Marpe, Heiko Schwarz, and Thomas Wiegand.
\newblock Context-based adaptive binary arithmetic coding in the h. 264/avc
  video compression standard.
\newblock {\em IEEE Transactions on circuits and systems for video technology},
  13(7):620--636, 2003.

\bibitem{mentzer2019practical}
Fabian Mentzer, Eirikur Agustsson, Michael Tschannen, Radu Timofte, and Luc~Van
  Gool.
\newblock Practical full resolution learned lossless image compression.
\newblock In {\em Proceedings of the IEEE Conference on Computer Vision and
  Pattern Recognition}, pages 10629--10638, 2019.

\bibitem{mentzer2020learning}
Fabian Mentzer, Luc~Van Gool, and Michael Tschannen.
\newblock Learning better lossless compression using lossy compression.
\newblock In {\em Proceedings of the IEEE/CVF Conference on Computer Vision and
  Pattern Recognition}, pages 6638--6647, 2020.

\bibitem{nalisnick2018deep}
Eric Nalisnick, Akihiro Matsukawa, Yee~Whye Teh, Dilan Gorur, and Balaji
  Lakshminarayanan.
\newblock Do deep generative models know what they don't know?
\newblock In {\em International Conference on Learning Representations}, 2019.

\bibitem{nalisnick2019detecting}
Eric Nalisnick, Akihiro Matsukawa, Yee~Whye Teh, and Balaji Lakshminarayanan.
\newblock Detecting out-of-distribution inputs to deep generative models using
  typicality.
\newblock {\em arXiv preprint arXiv:1906.02994}, 2019.

\bibitem{pennebaker1988overview}
William~B. Pennebaker, Joan~L. Mitchell, GG~Langdon, and Ronald~B Arps.
\newblock An overview of the basic principles of the q-coder adaptive binary
  arithmetic coder.
\newblock {\em IBM Journal of research and development}, 32(6):717--726, 1988.

\bibitem{rabbani2002jpeg2000}
Majid Rabbani.
\newblock Jpeg2000: Image compression fundamentals, standards and practice.
\newblock {\em Journal of Electronic Imaging}, 11(2):286, 2002.

\bibitem{roelofs1999png}
Greg Roelofs and Richard Koman.
\newblock {\em PNG: the definitive guide}.
\newblock O'Reilly \& Associates, Inc., 1999.

\bibitem{salimans2017pixelcnn++}
Tim Salimans, Andrej Karpathy, Xi~Chen, and Diederik~P. Kingma.
\newblock Pixelcnn++: Improving the pixelcnn with discretized logistic mixture
  likelihood and other modifications.
\newblock In {\em 5th International Conference on Learning Representations,
  {ICLR} 2017}.

\bibitem{schmidhuber1995predictive}
J{\"u}rgen Schmidhuber and Stefan Heil.
\newblock Predictive coding with neural nets: Application to text compression.
\newblock In {\em Advances in Neural Information Processing Systems}, pages
  1047--1054, 1995.

\bibitem{schmidhuber1996sequential}
J{\"u}rgen Schmidhuber and Stefan Heil.
\newblock Sequential neural text compression.
\newblock {\em IEEE Transactions on Neural Networks}, 7(1):142--146, 1996.

\bibitem{tatwawadi2018deepzip}
Kedar Tatwawadi.
\newblock Deepzip: Lossless compression using recurrent networks.
\newblock {\em URL https://web. stanford. edu/class/cs224n/reports/2761006.
  pdf}, 2018.

\bibitem{theis2015note}
Lucas Theis, A{\"{a}}ron van~den Oord, and Matthias Bethge.
\newblock A note on the evaluation of generative models.
\newblock In {\em 4th International Conference on Learning Representations,
  {ICLR} 2016}.

\bibitem{thomee2016yfcc100m}
Bart Thomee, David~A Shamma, Gerald Friedland, Benjamin Elizalde, Karl Ni,
  Douglas Poland, Damian Borth, and Li-Jia Li.
\newblock Yfcc100m: The new data in multimedia research.
\newblock {\em Communications of the ACM}, 59(2):64--73, 2016.

\bibitem{townsend2021lossless}
James Townsend.
\newblock {\em Lossless Compression with Latent Variable Models}.
\newblock PhD Thesis, UCL (University College London), 2021.

\bibitem{townsend2019practical}
James Townsend, Thomas Bird, and David Barber.
\newblock Practical lossless compression with latent variables using bits back
  coding.
\newblock In {\em International Conference on Learning Representations}, 2019.

\bibitem{townsend2019hilloc}
James Townsend, Thomas Bird, Julius Kunze, and David Barber.
\newblock Hilloc: lossless image compression with hierarchical latent variable
  models.
\newblock In {\em International Conference on Learning Representations}, 2020.

\bibitem{uria2013rnade}
Benigno Uria, Iain Murray, and Hugo Larochelle.
\newblock Rnade: The real-valued neural autoregressive density-estimator.
\newblock In {\em Proceedings of the 26th International Conference on Neural
  Information Processing Systems - Volume 2}, 2013.

\bibitem{berg2020idf++}
Rianne van~den Berg, Alexey~A Gritsenko, Mostafa Dehghani, Casper~Kaae
  S{\o}nderby, and Tim Salimans.
\newblock Idf++: Analyzing and improving integer discrete flows for lossless
  compression.
\newblock In {\em International Conference on Learning Representations}, 2020.

\bibitem{van2021overfitting}
Ties van Rozendaal, Iris A.~M. Huijben, and Taco Cohen.
\newblock Overfitting for fun and profit: Instance-adaptive data compression.
\newblock In {\em ICLR}, 2021.

\bibitem{wallace1990classification}
Chris~S Wallace.
\newblock Classification by minimum-message-length inference.
\newblock In {\em International Conference on Computing and Information}, pages
  72--81. Springer, 1990.

\bibitem{witten1987arithmetic}
Ian~H Witten, Radford~M Neal, and John~G Cleary.
\newblock Arithmetic coding for data compression.
\newblock {\em Communications of the ACM}, 30(6):520--540, 1987.

\bibitem{zhang2021ivpf}
Shifeng Zhang, Chen Zhang, Ning Kang, and Zhenguo Li.
\newblock ivpf: Numerical invertible volume preserving flow for efficient
  lossless compression.
\newblock In {\em Proceedings of the IEEE/CVF Conference on Computer Vision and
  Pattern Recognition}, pages 620--629, 2021.

\bibitem{ziv1977universal}
Jacob Ziv and Abraham Lempel.
\newblock A universal algorithm for sequential data compression.
\newblock {\em IEEE Transactions on information theory}, 23(3):337--343, 1977.

\end{thebibliography}
\bibliographystyle{plain}

\newpage

\appendix

\section{Appendix to Section 3}\label{app:formulation}
In this section, we adopt a bottom up explanation of OSOA and present two toy examples respectively for FIFO and FILO style entropy coders.
For the bottom up explanation part, we start from principles and properties of common entropy coders, briefly summarise the ideas to synergise DGMs with the entropy coders proposed in recent work and arrive at how OSOA accommodates DGMs based lossless compression algorithms with different styles of entropy coders.

\subsection{Entropy coders}\label{app:coders}
At the high level, the problem of lossless compression can be formulated as a problem finding a bijective map between the data domain and the code domain (the image of the data domain under such a map), such that the expected length of the variable in the code domain is less than the expected length of the variable in the data domain.
And a general principle for achieving shorter expected length is to assign a shorter code to a more frequent symbol.
Popular entropy coders, i.e., Huffman coding \cite{huffman1952method}, arithmetic coding \cite{witten1987arithmetic} and asymmetric numeral systems \cite{duda2013asymmetric}, define three different families of solutions to lossless compression under such principle.

\textbf{Toy Example 1} Here we use a simple example to explain the ideas and properties of above three entropy coders.
Assume the variable $x$ has possible values $\{a_1, a_2, a_3, a_4, a_5\}$, with probabilities in Tab \ref{tab:entropy_coder_simplex}.

\begin{table}[th!]
\caption{The simplex of the Toy Example 1}
\label{tab:entropy_coder_simplex}
\setlength{\tabcolsep}{3pt}
\begin{center}
\begin{small}
\begin{sc}
\begin{tabular}{lccccc}
\toprule
Symbol      & $a_1$  & $a_2$  & $a_3$  & $a_4$  & $a_5$ \\
\midrule
Probability & $0.32$ & $0.08$ & $0.16$ & $0.02$ & $0.42$\\
\bottomrule
\end{tabular}
\end{sc}
\end{small}
\end{center}
\vskip -0.1in
\end{table}

\begin{figure}[ht!]
\centering
\setlength{\tabcolsep}{0pt}
\begin{tabular}{l}
\includegraphics[width=0.45\textwidth]{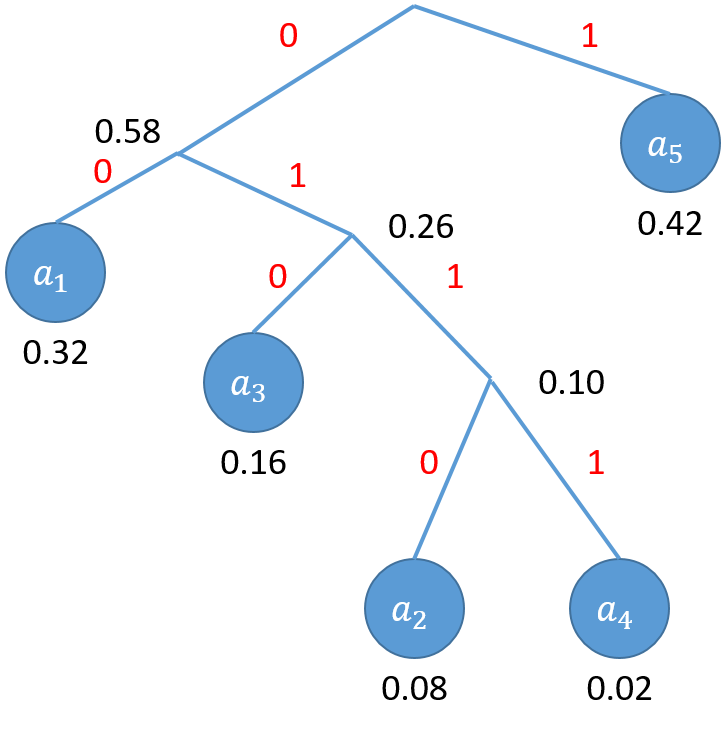}\\
\end{tabular}
\caption{The Huffman tree for Toy Example 1.}
\label{fig:huffman}
\end{figure}

\textbf{Huffman Coding}
Huffman coding gives such a bijective map through a binary tree generation process.
For a given probabilistic distribution, it starts from the two least probable symbols, $a_2$ and $a_4$, and forms a subtree $a_2 \bigvee a_4$, where the less probable symbol is the right leaf.
The two symbols $a_2$ and $a_4$ are substituted by the subtree $a_2\bigvee a_4$ with probability as sum of the two symbols, $0.10$.
Then the same iteration is conducted until no more nodes can be added, cf. Fig \ref{fig:huffman}.
For a certain symbol, the code of it reads from the root to the leaf node, e.g., $0110$ for $a_2$.
And the decoding process is to look up the codebook given by the tree.
Note that Huffman coding is a prefix code, in the sense that no codeword is the prefix of another code.
As a result, codewords of different symbols are isolated and a symbol sequence can be decode from the codeword sequence as the original order.
For example, the code word for the symbol sequence $a_5a_3a_2$ is $10100110$.

\begin{figure}[ht!]
\centering
\setlength{\tabcolsep}{0pt}
\begin{tabular}{l}
\includegraphics[width=0.6\textwidth]{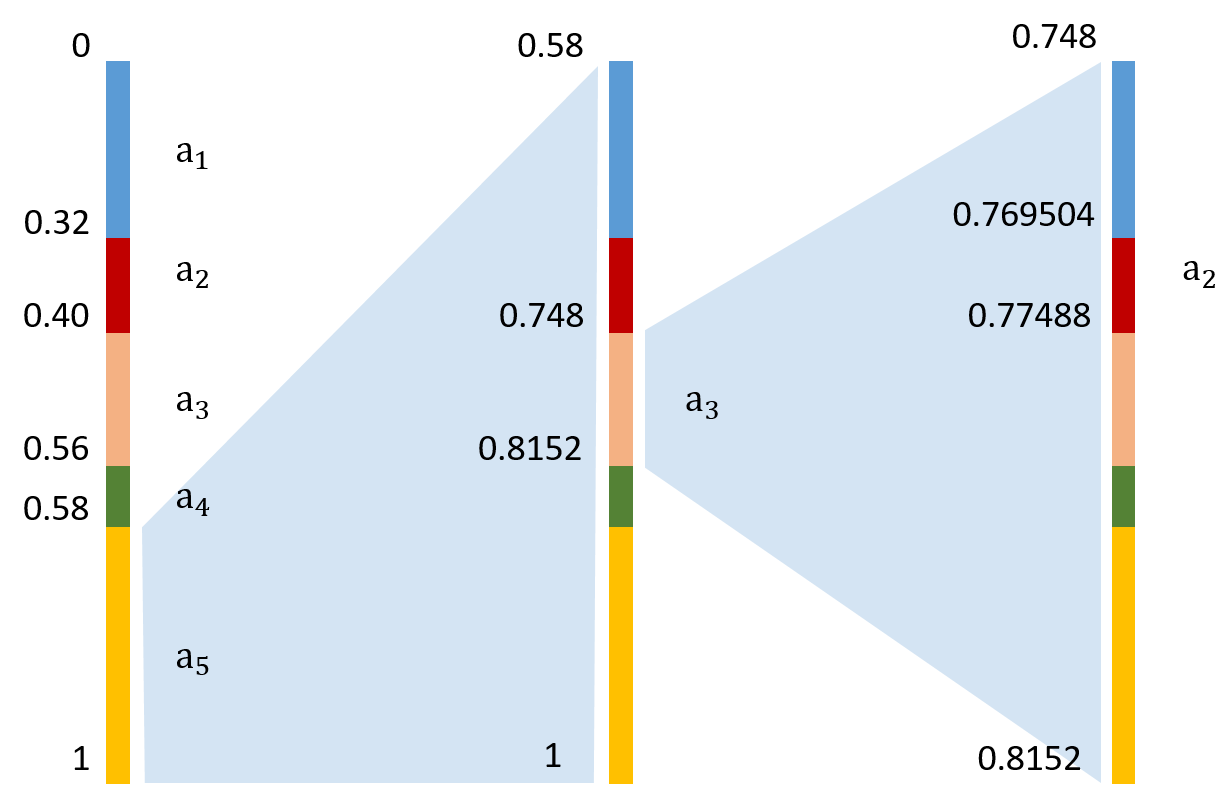}\\
\end{tabular}
\caption{Arithmetic coding for Toy Example 1.}
\label{fig:ac}
\end{figure}

\textbf{Arithmetic Coding} 
Arithmetic coders partition an interval into subintervals according to a simplex, associate one symbol with one subinterval and use a real number in that interval to represent the associated symbol.
To encode a symbol sequence, arithmetic coders start from the unit interval $I_0 = [0,1]$ and find the associated subinterval $I_1$ for the first symbol.
Then $I_1$ will be partitioned again according to the simplex and a subinterval $I_2$ will be found for the second symbol.
The iterative process is conducted until the end of the sequence and a real number in the last interval is used to represent the symbol sequence.
To decode a sequence, one just need to examine which subinterval in $I_0$ the real number belongs to and decodes the associated symbol as the first one.
Then one examines which subinterval in $I_1$ the real number belongs to and decodes the second symbol.
To enable the iterative process to stop at the correct step, a termination symbol can be introduced.
Without loss of generality, we assume $a_2$ is the termination symbol, where the codeword for $a_5a_3a_2$ by AC can be $0.77$, cf. Fig. \ref{fig:ac}.
Note that the full sequence $a_5a_3a_2$ is represented by a single codeword and one has to decode $a_5$ before being able to decode $a_3$.
As a result, entropy coders of such fashion are referred to as of First-In-First-Out (FIFO) stytle, i.e., the queue style,.

\textbf{Asymmetric Numeral Systems}
Here we present rANS.
rANS first approximates the probability of each symbol $a_i$ with a rational number $\frac{\ell_{a_i}}{m}$, where $m=\sum \ell_{a_i}$.
The cumulative count is defined as $b_{a_i}=\sum_{j=1}^{i-1}\ell_{a_j}$.
Furthermore, the inverse cumulative count function is defined as
\begin{equation}
    b^{-1}(y) = \arg\min_{a_j} \{y<\sum_{i=1}^j \ell_{a_i}\}.
\end{equation}
For a sequence $\{s_t\}_{t=1}^T$, the function to encode $s_t$ is
\begin{equation}
    x_{t} = \lfloor \frac{x_{t-1}}{\ell_{s_{t}}} \rfloor * m + \texttt{mod}(x_{t-1}, \ell_{s_{t}}) + b_{s_{t}}.
\end{equation}
And the function to decode $s_t$ is
\begin{align}
s_{t} &= b^{-1}(\texttt{mod}(x_{t}, m)),\\
x_{t-1} &= \lfloor \frac{x_{t}}{m} \rfloor * \ell_{s_{t}} + \texttt{mod} (x_{t}, m) - b_{s_{t}}.
\end{align}
Note that $\{x_t\}_{t=0}^T$ is the codeword sequence generated during the encoding process and only $x_T$ is needed for decoding.
The initial codeword state $x_0$ can be initialised as $0$.

Using the same Toy Example 1, we first compute the information as listed in Tab \ref{tab:rans_simplex}.
\begin{table}[th!]
\caption{Information needed of rANS for Toy Example 1}
\label{tab:rans_simplex}
\setlength{\tabcolsep}{3pt}
\begin{center}
\begin{small}
\begin{sc}
\begin{tabular}{lccccc}
\toprule
Symbol      & $a_1$  & $a_2$  & $a_3$  & $a_4$  & $a_5$ \\
\midrule
Probability & $0.32$ & $0.08$ & $0.16$ & $0.02$ & $0.42$\\
$\ell_{a_i}$& $32$   & $8$    & $16$   & $2$    & $42$\\
$b_{a_i}$   & $0$    & $32$   & $40$   & $56$   & $58$\\
\bottomrule
\end{tabular}
\end{sc}
\end{small}
\end{center}
\vskip -0.1in
\end{table}

The encoding process for the sequence $a_5a_3a_2$ is as follows
\begin{enumerate}
    \item $x_0 = 0$
    \item $x_1 = \lfloor \frac{0}{42} \rfloor * 100 + \texttt{mod}(0, 42) + 58 = 58$
    \item $x_2 = \lfloor \frac{58}{16} \rfloor * 100 + \texttt{mod}(58, 16) + 40 = 350$
    \item $x_3 = \lfloor \frac{350}{8} \rfloor * 100 + \texttt{mod}(350, 8) + 32 = 4338$.
\end{enumerate}
And the codeword for the sequence is $4338$.

The decoding process is as follows
\begin{enumerate}
    \item $x_3 = 4338$
    \item $s_3 = b^{-1}(\texttt{mod}(4338, 100)) = a_2$ and $x_2 = \lfloor \frac{4338}{100} \rfloor * 8 + \texttt{mod} (4338, 100) - 32 = 350$
    \item $s_2 = b^{-1}(\texttt{mod}(350, 100)) = a_3$ and $x_1 = \lfloor \frac{350}{100} \rfloor * 16 + \texttt{mod} (350, 100) - 40 = 58$
    \item $s_1 = b^{-1}(\texttt{mod}(58, 100)) = a_5$ and $x_0 = \lfloor \frac{58}{100} \rfloor * 42 + \texttt{mod} (58, 100) - 58 = 0$.
\end{enumerate}
Different from Huffman coding and Arithmetic Coding, ANS decodes the symbols in the reverse order as the encoding one.
As a result, the fashion of coding with ANS is referred to as of First-In-Last-Out (FILO) style, i.e., the stack style.

\textbf{Adaptive coding}
We presented three coding examples respectively for each entropy coder with a fixed distribution of the symbols.
All of the mentioned entropy coders allow for changes of the distributions during the coding process.
For Huffman coding, it involves dynamic adjustments of the tree \cite{knuth1985dynamic}.
For AC, it involves different partitions at different steps \cite{marpe2003context}.
For ANS, it involves different values of $\ell_{a_i}$'s and $b_{a_i}$'s \cite{duda2013asymmetric}.
While current DGMs based lossless compression algorithms only consider static coding, OSOA introduces and validates adaptive coding for DGMs based lossless compression algorithms.

\subsection{DGMs based lossless compression}\label{app:dgm}
The entropy coders reviewed in Sec \ref{app:coders} are elementary functional units in a DGMs based lossless compression algorithm. 
As reviewed in Sec. 2, recent research focus on how to synergise above entropy coders with different types of DGMs, i.e., VAEs, Normalizing Flows, etc.
The main reason is that for a data variable $\mathbf{x}$, normally we do not have direct access to the explicit form of $p(\mathbf{x})$ but instead some factorisation form of it in DGMs.
Here we use a single latent VAE as an example to illustrate the components in a DGM based lossless compression algorithm, i.e., a deep generative model and an entropy coder and an algorithm to connect the model and the coder.

\textbf{Toy Example 2}
In a single latent VAE, we have the observation variable $\mathbf{x}$ and the latent variable $\mathbf{z}$.
The modelled distributions are prior $p(\mathbf{z})$, likelihood $p(\mathbf{x}|\mathbf{z})$ and the approximate posterior $q(\mathbf{z}|\mathbf{x})$.
For a chosen discretisation scheme, we denote the discrete distributions by $\bar{p}(\mathbf{z})$, $\bar{p}(\mathbf{x}|\mathbf{z})$ and $\bar{q}(\mathbf{z}|\mathbf{x})$.
The entropy coder adopted here is rANS.
The algorithm connects the VAE and rANS is called bits back ANS (bb-ANS) \cite{townsend2019practical}.
In bb-ANS, an auxiliary amount of initial bits are required, which is denoted by $c_0$.

To encode a given sample $\mathbf{x}_1$, one 1) decodes $\mathbf{z}_1$ from $c_0$ with $\bar{q}(\mathbf{z}|\mathbf{x})$ using rANS and gets the code $c_0^1$, 2) encodes $\mathbf{x}_1$ to $c_0^1$ with $\bar{p}(\mathbf{x}|\mathbf{z})$ using rANS and gets the code $c_0^2$ and 3) encodes $\mathbf{z}_1$ to $c_0^2$ with $\bar{p}(\mathbf{z})$ and gets $c_1$.

To decode $\mathbf{x}_1$ from $c_1$, since ANS is an FILO style coder, one has to reverse the above process and swap the operations of encoding and decoding.
Specifically, one 1) decodes $\mathbf{z}_1$ from $c_1$ with $\bar{p}(\mathbf{z})$ using rANS and gets $c_0^2$, 2) decodes $\mathbf{x}_1$ from $c_0^2$ with $\bar{p}(\mathbf{x}|\mathbf{z})$ using rANS and gets $c_0^1$ and 3) encodes $\mathbf{z}_1$ to $c_0^1$ with $\bar{q}(\mathbf{z}|\mathbf{x})$ using rANS and gets $c_0$.

Due to the same reason of FILO coders, if one first encodes $\mathbf{x}_1$ with above algorithm and then encodes $\mathbf{x}_2$ with above algorithm, one can only first decode $\mathbf{x}_2$ and then decodes $\mathbf{x}_1$.
Note that for FIFO coders, e.g., AC, one can only first decode $\mathbf{x}_1$ and then decode $\mathbf{x}_2$.

\subsection{OSOA}
From Sec. \ref{app:dgm}, one can see that a DGM based lossless compression framework adds one more layer, i.e., the deep generative model layer, to the entropy coder layer and synergise these two layers by an algorithm, e.g., bb-ANS in the above example.
The proposed OSOA actually adds another layer, i.e., the model adaptation layer, to DGMs based lossless compression frameworks and synergise these two layers by different algorithms for FIFO style and FILO style entropy coders, which is adopted in a particular DGM based lossless compression framework.
In this section, we present two simple examples respectively for DGMs based lossless compression frameworks with FIFO entropy coders and FILO entropy coders.

\textbf{Toy Example 3}
We have three batches to compress, i.e., $B_1$, $B_2$ and $B_3$ and a DGM based lossless compression framework with a FIFO entropy coder.
Moreover, we have a pretrained model $p_0$ for the DGM in the above framework.

\begin{figure}[ht!]
\centering
\setlength{\tabcolsep}{0pt}
\begin{tabular}{l}
\includegraphics[width=0.45\textwidth]{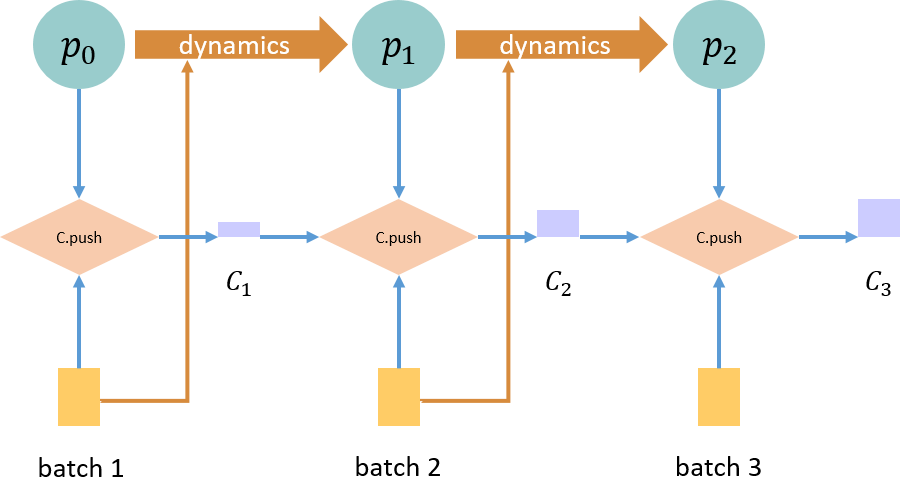}\\
\end{tabular}
\caption{An illustration of OSOA encoding with arithmetic coding (AC) for Toy Example 3. C.push denotes the encoding operation of the AC codec.}
\label{fig:ac_encoding_app}
\end{figure}

The encoding include three steps as follows and illustrated in Fig. \ref{fig:ac_encoding_app}.
\begin{enumerate}
    \item Use $p_0$ to compress $B_1$ with the inbuilt algorithm of the framework, then use $B_1$ to update $p_0$ and get $p_1$.
    \item Use $p_1$ to compress $B_2$ with the inbuilt algorithm of the framework, then use $B_2$ to update $p_1$ and get $p_2$.
    \item Use $p_2$ to compress $B_3$ with the inbuilt algorithm of the framework.
\end{enumerate}
Note that since we do not have more batches, we do not need to use $B_3$ to update $p_2$.

\begin{figure}[ht!]
\centering
\setlength{\tabcolsep}{0pt}
\begin{tabular}{l}
\includegraphics[width=0.45\textwidth]{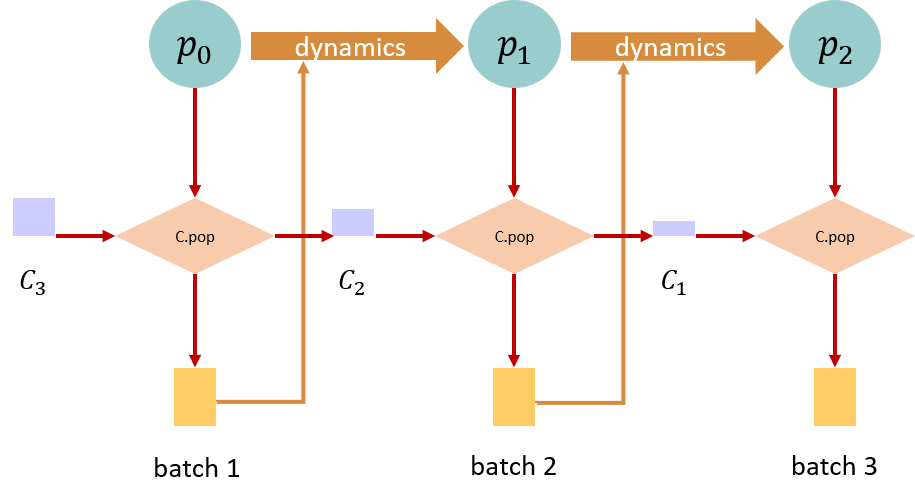}\\
\end{tabular}
\caption{An illustration of OSOA decoding with arithmetic coding (AC) for Toy Example 3. C.pop denotes the decoding operation of the AC codec.}
\label{fig:ac_decoding_app}
\end{figure}

The decoding include three steps as follows and illustrated in Fig. \ref{fig:ac_decoding_app}.
\begin{enumerate}
    \item Use $p_0$ to decompress $B_1$ with the inbuilt algorithm of the framework, then use $B_1$ to update $p_0$ and get $p_1$.
    \item Use $p_1$ to decompress $B_2$ with the inbuilt algorithm of the framework, then use $B_2$ to update $p_1$ and get $p_2$.
    \item Use $p_2$ to decompress $B_3$ with the inbuilt algorithm of the framework.
\end{enumerate}

\textbf{Toy Example 4}
We have six batches to compress, i.e., $B_1$, $B_2$, $B_3$, $B_4$, $B_5$ and $B_6$ and a DGM based lossless compression framework with a FILO entropy coder.
Moreover, we have a pretrained model $p_0$ for the DGM in the above framework.
Here we divide the six batches into two chunks with consecutive three batches per chunk.

\begin{figure*}[ht!]
\centering
\setlength{\tabcolsep}{0pt}
\begin{tabular}{l}
\includegraphics[width=0.95\textwidth]{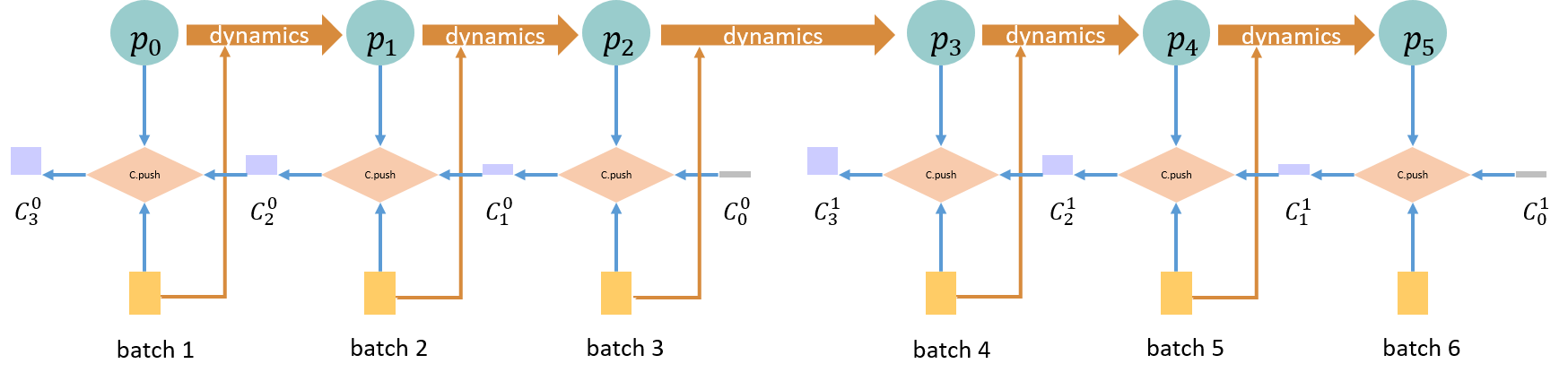}\\
\end{tabular}
\caption{An illustration of OSOA encoding with bits back asymmetric numerical system (bb-ANS) for Toy Example 4. C.push denotes the encoding operation of the ANS codec.}
\label{fig:ans_encoding_app}
\end{figure*}

The encoding include six steps as follows and illustrated in Fig. \ref{fig:ans_encoding_app}.
\begin{enumerate}
    \item Use $p_0$ to evaluate the pmf needed to compress $B_1$ with the inbuilt algorithm of the framework and add the pmf and batch to cache, then use $B_1$ to update $p_0$ and get $p_1$.
    \item Use $p_1$ to evaluate the pmf needed to compress $B_2$ with the inbuilt algorithm of the framework and add the pmf and batch to cache, then use $B_2$ to update $p_1$ and get $p_2$.
    \item Use $p_2$ to evaluate the pmf needed to compress $B_3$ with the inbuilt algorithm of the framework and add the pmf and batch to cache, then use $B_3$ to update $p_2$ and get $p_3$.
          Start an independent process to consecutively compress $B_3$, $B_2$ and $B_1$ with the information in the cache. Clear the cache for the first three batches.
    \item Use $p_3$ to evaluate the pmf needed to compress $B_4$ with the inbuilt algorithm of the framework and add the pmf and batch to cache, then use $B_4$ to update $p_3$ and get $p_4$.
    \item Use $p_4$ to evaluate the pmf needed to compress $B_5$ with the inbuilt algorithm of the framework and add the pmf and batch to cache, then use $B_5$ to update $p_4$ and get $p_5$.
    \item Use $p_5$ to evaluate the pmf needed to compress $B_6$ with the inbuilt algorithm of the framework and add the pmf and batch to cache.
          Start an independent process to consecutively compress $B_6$, $B_5$ and $B_4$ with the information in the cache. Clear the cache for the last three batches.
\end{enumerate}
Note that since we do not have more batches, we do not need to use $B_6$ to update $p_5$.

\begin{figure*}[ht!]
\centering
\setlength{\tabcolsep}{0pt}
\begin{tabular}{l}
\includegraphics[width=0.95\textwidth]{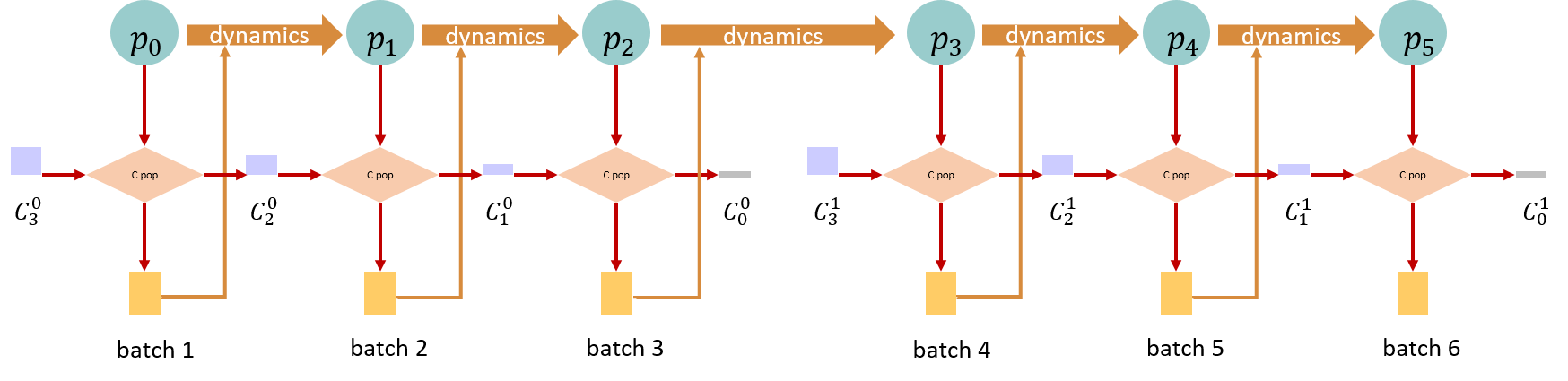}\\
\end{tabular}
\caption{An illustration of OSOA decoding with bits back asymmetric numerical system (bb-ANS) for Toy Example 4. C.pop denotes the decoding operation of the ANS codec.}
\label{fig:ans_decoding_app}
\end{figure*}

The decoding include six steps as follows and illustrated in Fig. \ref{fig:ans_decoding_app}.
\begin{enumerate}
    \item Use $p_0$ to decompress $B_1$ with the inbuilt algorithm of the framework, then use $B_1$ to update $p_0$ and get $p_1$.
    \item Use $p_1$ to decompress $B_2$ with the inbuilt algorithm of the framework, then use $B_2$ to update $p_1$ and get $p_2$.
    \item Use $p_2$ to decompress $B_3$ with the inbuilt algorithm of the framework, then use $B_3$ to update $p_2$ and get $p_3$.
    \item Use $p_3$ to decompress $B_4$ with the inbuilt algorithm of the framework, then use $B_4$ to update $p_3$ and get $p_4$.
    \item Use $p_4$ to decompress $B_5$ with the inbuilt algorithm of the framework, then use $B_5$ to update $p_4$ and get $p_5$.
    \item Use $p_5$ to decompress $B_6$ with the inbuilt algorithm of the framework.
\end{enumerate}
Note that since we do not have more batches, we do not need to use $B_6$ to update $p_5$.

\textbf{Vanilla OSOA}
In vanilla OSOA, the steps of using $B_i$ to update $p_{i-1}$ to $p_i$ will be using a gradient based optimiser to update $p_{i-1}$ to $p_i$ with the gradient evaluated on $B_i$, in both OSOA Encoding and OSOA Decoding.

\section{Appendix to Section 4}\label{app:experiments}

\subsection{Samples}
\begin{figure*}[ht!]
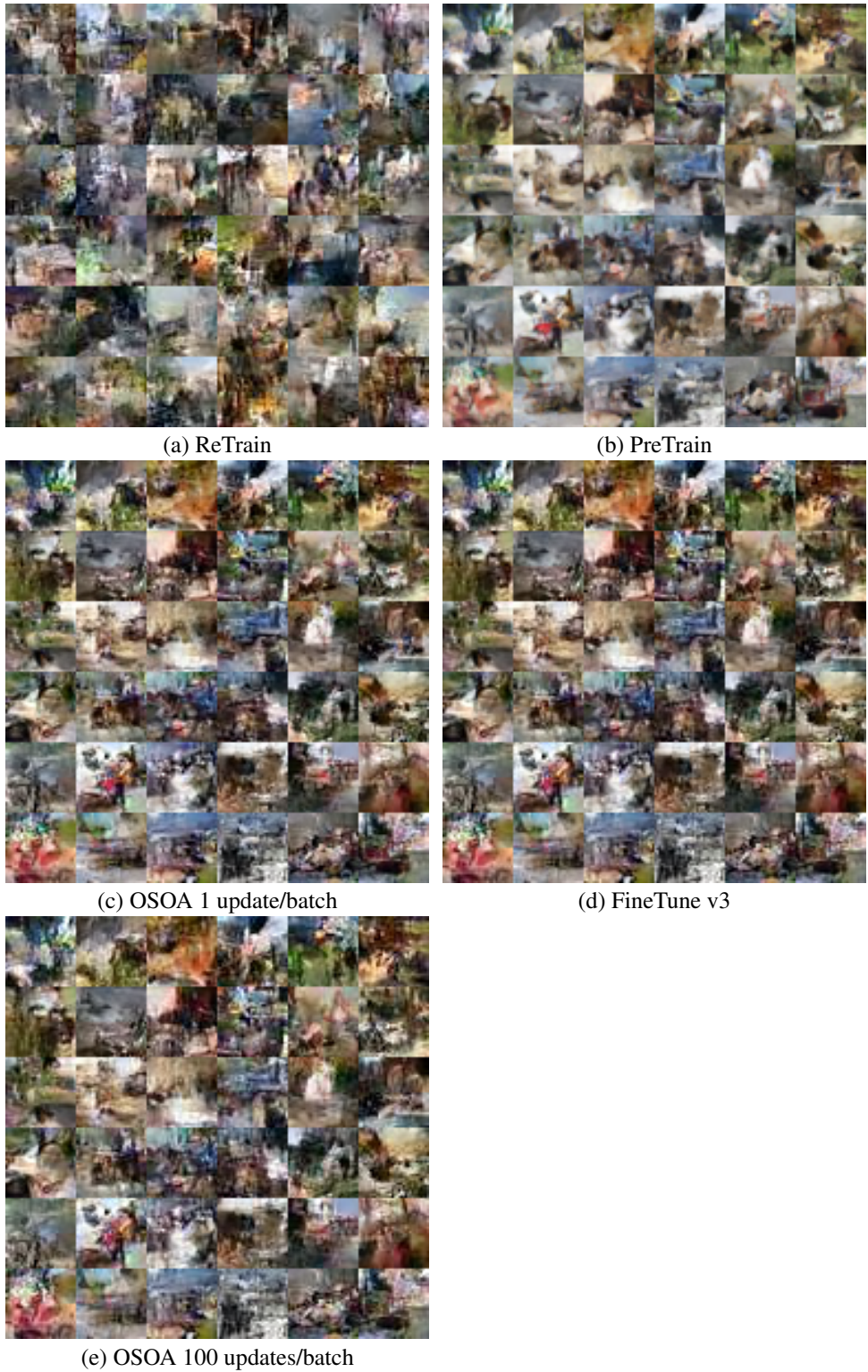

\centering
\setlength{\tabcolsep}{3pt}
\begin{tabular}{cc}
\includegraphics[width=0.45\textwidth]{fig/hilloc_set32_retrain.png} &
\includegraphics[width=0.45\textwidth]{fig/hilloc_cifar10_pretrain.png} \\
(a) ReTrain & (b) PreTrain\\
\includegraphics[width=0.45\textwidth]{fig/hilloc_osoa_1_step_batch.png} & 
\includegraphics[width=0.45\textwidth]{fig/hilloc_fine_tune_v3.png}  \\
(c) OSOA 1 update/batch & (d) FineTune v3\\
\includegraphics[width=0.45\textwidth]{fig/hilloc_osoa_100_step_batch.png} &\\
 (e) OSOA 100 updates/batch &\\
\end{tabular}
\caption{36 images sampled from (a) ReTrain: HiLLoC trained on SET32 from scratch, (b) PreTrain: CIFAR10 pretrained HiLLoC, (d) FineTune v3: fine-tuning CIFAR10 pretrained HiLLoC for 20 epochs on SET32, (c) and (e) OSOA 1 and 100 updates/batch: the final checkpoint of vanilla OSOA with 1 and 100 update steps per batch from CIFAR10 pretrained HiLLoC, respectively.}
\label{fig:imgs_app}
\end{figure*}

Please find Fig \ref{fig:imgs_app} for visually larger images of those shown in Fig 3 in the main file.

\subsection{Existing assets}
Existing assets used in this work include existing data and existing software.

\textbf{Data} 
Existing datasets used in this work, CIFAR10, ImageNet32 and YFCC100m, are public datasets freely used for research purpose.
Both CIFAR10 and ImageNet32 are of the MIT license and YFCC100m is of Creative Commons licenses.
SET32/64/128 are sampled and subsampled from images in YFCC100m as expained in Sec. 4.1.
Since CIFAR10, ImageNet32 and SET32 are of very low resolution, images from these datasets can be regarded with almost no personally identifiable information.
SET64 and SET128 are of higher resolutions than $32\times 32$ and may include personal identifiable information, but we are not showing direct samples from them nor samples from models trained on them.
Further, we discuss the importance of information protection with DGMs based lossless compression algorithms in Sec. 4.3 with samples of low resolution.

\textbf{Software}
The open source software for deep learning and data preprocessing are included and cited in Sec. 4.1.
Here we add more details on the code for deep generative models and associated lossless compression.
The code for the RVAE model adopted in HiLLoC and the IAF RVAE model shown in this work are adapted from the official code release of the work \cite{kingma2016iaf} and the official code release of the work \cite{townsend2019hilloc} both with the MIT license.
The codec used with HiLLoC is the package Craystack released with \cite{townsend2019hilloc} with the MIT license.
The model and the codec used with IDF++ is released with IDF~\cite{hoogeboom2019integer} in which slight modifications are performs to change IDF to IDF++. The codes are with MIT license.

\subsection{Experiments resources}
The infrastructure for experiments of HiLLoC is Intel(R) Xeon(R) CPU @ 2.60GHz$\times 16$ CPU with an Nvidia V100 32GB GPU. The CPU used in IDF++ is the same as that in HiLLoC while the GPU is Nvidia P100 16GB GPU.

\subsection{Training details}
\textbf{HiLLoC pretraining}
For the RVAE model (HiLLoC), we adopt the same architecture as the one used in the HiLLoC paper, as specified in Sec. 4.1.
We use the amount of free bits $0.1$, learning rate $0.002$ and batch size $16$ for pretraining on CIFAR10 and ImageNet32.
For the IAF RVAE model, we adopt the same configuration as RVAE with additionally enabling IAF for the approximate posterior.
We use the amount of free bits $0.1$, learning rate $0.002$ and batch size $32$ for pretraining on CIFAR10 and ImageNet32.
HiLLoC has $40998823$ trainable float32 parameters and IAF RAVE has $49864615$ trainable float32 parameters.
All of the pretrained models are pretrained for more than $60$ hours.

\textbf{HiLLoC OSOA}
For OSOA in HiLLoC and IAF RVAE, we use the random seed 14865 and learning rate at $0.0002$ for all experiments unless otherwise specified.
Since there are $2^{17}$ images in total for SET32/64/128, the vanilla OSOA without early stopping is conducted for $512$, $2048$ and $8192$ steps respectively for SET32/64/128.

\textbf{IDF++ pretraining}
For the IDF++ model, we adopt the same architecture as that in IDF++ paper. For pretraining, we use almost the same hyper-parameter used in IDF \cite{hoogeboom2019integer}. IDF++ has about $56.8M$ trainable parameters.

\textbf{IDF++ OSOA} For OSOA in IDF++, we use the random seed 5 and learning rate at $0.0003$ for all experiments unless otherwise specified. 

\subsection{Performance by changing the hyperparameters}
\begin{figure}[ht!]
\centering
\setlength{\tabcolsep}{0pt}
\begin{tabular}{cc}
\includegraphics[width=0.40\textwidth]{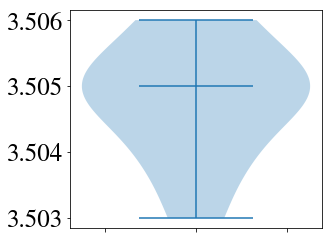} &
\includegraphics[width=0.40\textwidth]{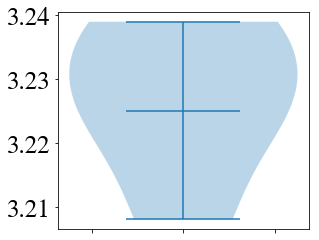}\\
\end{tabular}
\caption{Violin plots of OSOA with different random seeds shown with CIFAR10 pretrained HiLLoC (left) and CIFAR10 pretrained IDF++ (right).}
\label{fig:seed}
\end{figure}

\textbf{Error bars with different random seeds}
To evaluate the sensitivity of vanilla OSOA with respective to random seeds, we randomly choose 10 different random seeds for CIFAR10 pretrained HiLLoC and 6 random seeds for CIFAR10 pretrained IDF++ on SET32, and show the violin plots in Fig. \ref{fig:seed}.
One can see that OSOA admits performances with plausible consistency of different random seeds.

\begin{figure}[ht!]
\centering
\setlength{\tabcolsep}{0pt}
\begin{tabular}{cc}
\includegraphics[width=0.40\textwidth]{fig/32_lr.png} &
\includegraphics[width=0.40\textwidth]{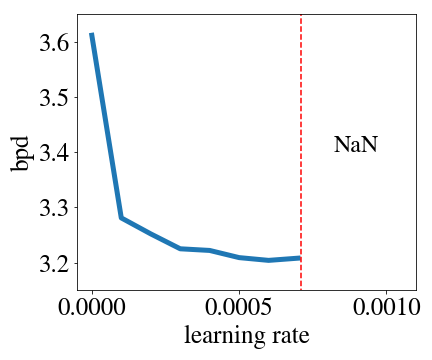}\\
\end{tabular}
\caption{The learning rate smile: bpd of OSOA with HiLLoC (left) and IDF++ (right) of different learning rates on SET32. "NaN" denotes the failure of training with larger learning rates.}
\label{fig:lr}
\end{figure}

\textbf{OSOA with different learning rates}
Learning rate is one of the most important factors for fine-tuning and a proper learning rate is usually more desirable.
Here we line-searched the interval $[0.0001, 0.012]$ for OSOA with HiLLoC and $[0.0001, 0.001]$ with IDF++, and showcase the learning rate smile curve in Fig. \ref{fig:lr}.
The random seed for HiLLoC is fixed at $14865$ and the random seed for IDF++ is fixed at $5$.
We find for HiLLoC, while the pre-training learning rate is $0.002$, a slightly larger learning rate $0.003$ can achieve better OSOA performance. For IDF++, a smaller learning rate is preferred, while larger ones lead to the failure of training, which highlights the importance of a proper learning rate choice.
Moreover, one can see in Fig. \ref{fig:lr} that even with the learning rate difference at the order of magnitude of $0.001$, a difference of bpd at the magnitude of $0.1$ can be witnessed.

\subsection{Real bpd values}
Note that the theoretical bpd values reported in this paper are evaluated based on the discretised distributions of models for coding.
Depending on the design and implementation of the codec, the real bpd value will be slightly higher than the theoretical one, e.g., an extra cost of less than 32 bits \cite{townsend2019hilloc} for bits-back ANS, which is negligible for large size datasets \cite{townsend2019hilloc, berg2020idf++}.
Since the comparison between OSOA and baselines are for the same model and same codec, theoretical bpd results are sufficient for comparison.
The codec for HiLLoC is Craystack \cite{townsend2019hilloc}, which is a prototype purpose python implemented codec developed in the work \cite{townsend2019hilloc} and for IDF++ is the self-implemented AC coder.
The real bpd values of HiLLoC and IDF++ on SET32 are shown in Tab \ref{tab:real_bpd}.
One can see that the differences between the theoretical bpd values and the real bpd values for OSOA and FineTune baselines are at the order of magnitude of $0.01$ for HiLLoC and the order of magnitude of $0.001$ for IDF++.
Further, the conclusion of the comparison between the space efficiency of OSOA and FineTune baselines holds.

\begin{table}[th!]
\caption{Real bpd values}
\label{tab:real_bpd}
\setlength{\tabcolsep}{3pt}
\begin{center}
\begin{small}
\begin{sc}
\begin{tabular}{lcccccc}
\toprule
                 & SET & PreTrain & OSOA  & FineTune v1        & FineTune v2        & FineTune v3\\
\midrule
HiLLoC (CIFAR10) & 32  &  4.025   & 3.565 & 3.421 \FG{(0.144)} & 3.364 \FG{(0.201)} & 3.257 \FG{(0.308)}    \\
\midrule
IDF++ (CIFAR10)  & 32  &  3.614    & 3.227 & 3.092 \FG{(0.135)} & 3.047 \FG{(0.180)} & 2.837 \FG{(0.390)}  \\
\bottomrule
\end{tabular}
\end{sc}
\end{small}
\end{center}
\vskip -0.1in
\end{table}

\end{document}